\documentclass{article}

\PassOptionsToPackage{table}{xcolor}
\PassOptionsToPackage{numbers,compress}{natbib}

\usepackage[preprint]{neurips_2025}

\usepackage[utf8]{inputenc} 
\usepackage[T1]{fontenc}    
\usepackage{hyperref}       
\usepackage{booktabs}       
\usepackage{amsmath}        
\usepackage{nicefrac}       
\usepackage{microtype}      

\usepackage{enumitem}
\usepackage{graphicx}
\usepackage{multirow}
\usepackage{amsfonts}       
\usepackage{amssymb}        
\usepackage{tabularx}
\usepackage{pifont}
\usepackage{tikz}
\usepackage{tcolorbox}
\usepackage{float}
\usepackage{makecell}
\newcommand{\cmark}{\ding{51}}  
\usepackage{authblk}

\usepackage{titletoc} 

\title{KG-TRACES: Enhancing Large Language Models with Knowledge Graph-constrained Trajectory Reasoning and Attribution Supervision}

\author{
Rong Wu$^{1,2}$, Pinlong Cai$^{2}$, Jianbiao Mei$^{1,2}$, Licheng Wen$^2$, Tao Hu$^{2,3}$, \\Xuemeng Yang$^{2}$, Daocheng Fu$^{2,4}$, Botian Shi$^{2,}$\thanks{corresponding author} \vspace{7pt} \\

$^{1}$ Zhejiang University
$^{2}$ Shanghai Artificial Intelligence Laboratory \\
$^{3}$ University of Science and Technology of China
$^{4}$ Fudan University
}

\begin{document}

\maketitle

\begin{abstract}
Large language models (LLMs) have made remarkable strides in various natural language processing tasks, but their performance on complex reasoning problems remains hindered by a lack of explainability and trustworthiness. This issue, often manifesting as hallucinations or unattributable reasoning processes, limits their applicability in complex reasoning scenarios. To address this, we propose \textbf{K}nowledge \textbf{G}raph-constrained \textbf{T}rajectory \textbf{R}easoning \textbf{A}ttribution and \textbf{C}hain \textbf{E}xplanation \textbf{S}upervision (\textbf{KG-TRACES}), a novel framework that enhances the reasoning ability of LLMs through explicit supervision over reasoning paths and processes. KG-TRACES jointly supervises the model to: (1) predict symbolic relation paths, (2) predict full triple-level reasoning paths, and (3) generate attribution-aware reasoning processes grounded in the reasoning paths. At inference phase, the model adapts to both KG-available and KG-unavailable scenarios, retrieving reasoning paths from a KG when possible or predicting plausible reasoning paths with only intrinsic knowledge when not. This design enables the model to reason in an explainable and source-attributable pattern. Through extensive experiments on complex reasoning tasks, we demonstrate that KG-TRACES significantly outperforms existing SOTA: it improves Hits@1 by 1.6\% and F1 by 4.7\% on WebQSP, and achieves improvements of 4.8\% in Hits@1 and 2.1\% in F1 on CWQ. Moreover, we show its transferability to specialized domains such as medicine. By visualizing the intermediate steps of reasoning processes, we further analyze the intermediate steps of the reasoning processes, which become increasingly consistent with the final conclusion. Code is available at \textcolor{magenta}{\url{https://github.com/Edaizi/KG-TRACES}}.

\end{abstract}

\section{Introduction}
\label{intro}
Large language models (LLMs) have achieved remarkable success across a wide range of natural language processing tasks. Yet, their performance on complex multi-hop reasoning remains hindered by a lack of explainability and attribution~\cite{huang2025survey, Li2024TheDAA}. In particular, current models often generate hallucinated intermediate steps or ungrounded conclusions, which severely limits their applicability in domains that demand explainable and faithful reasoning, such as open-domain question answering, scientific discovery or clinical decision support~\cite{ji-etal-2023-towards, sahoo-etal-2024-comprehensive}.

Recent advances have explored several directions to mitigate these issues. Chain-of-Thought (CoT) prompting encourages step-by-step reasoning by examples~\cite{kojima2022large, li2025structured, wang-etal-2023-cue}, while retrieval-augmented generation (RAG) methods attempt to ground model outputs with external knowledge~\cite{lewis2020retrieval, jiang2023active, xiong2024benchmarking}. Knowledge graph (KG) capture abundant factual world knowledge, organize the structural relationships between entities from fragmented information and store knowledge in the form of triples. Researchers have developed KG-enhanced LLMs approaches (which augmented LLMs with structured factual knowledge from KG to reasoning) attempting to mitigate LLM's hallucination and lack of faithful knowledge~\cite{wang2023knowledge, jiang2023unikgqa, wen-etal-2024-mindmap, ma2024think}. Nevertheless, these methods either only rely on loose prompt-based constraints with external KG information~\cite{wen-etal-2024-mindmap, sun2024thinkongraph}, or depend heavily on external knowledge retrieval, suffering significant performance degradation under limited KG access~\cite{yasunaga-etal-2021-qa, luo2024reasoning}. Moreover, while they incorporate structured and factual KG information, they fail to provide attributable reasoning processes or explain the sources of their conclusions. This lack of reasoning attribution fundamentally undermines the goal of improving trustworthiness, as it fails to align with the very premise of using structured world knowledge to enhance reasoning explainable and attributable reasoning.

\begin{figure}[!t]
    \centering
    \includegraphics[width=1\linewidth]{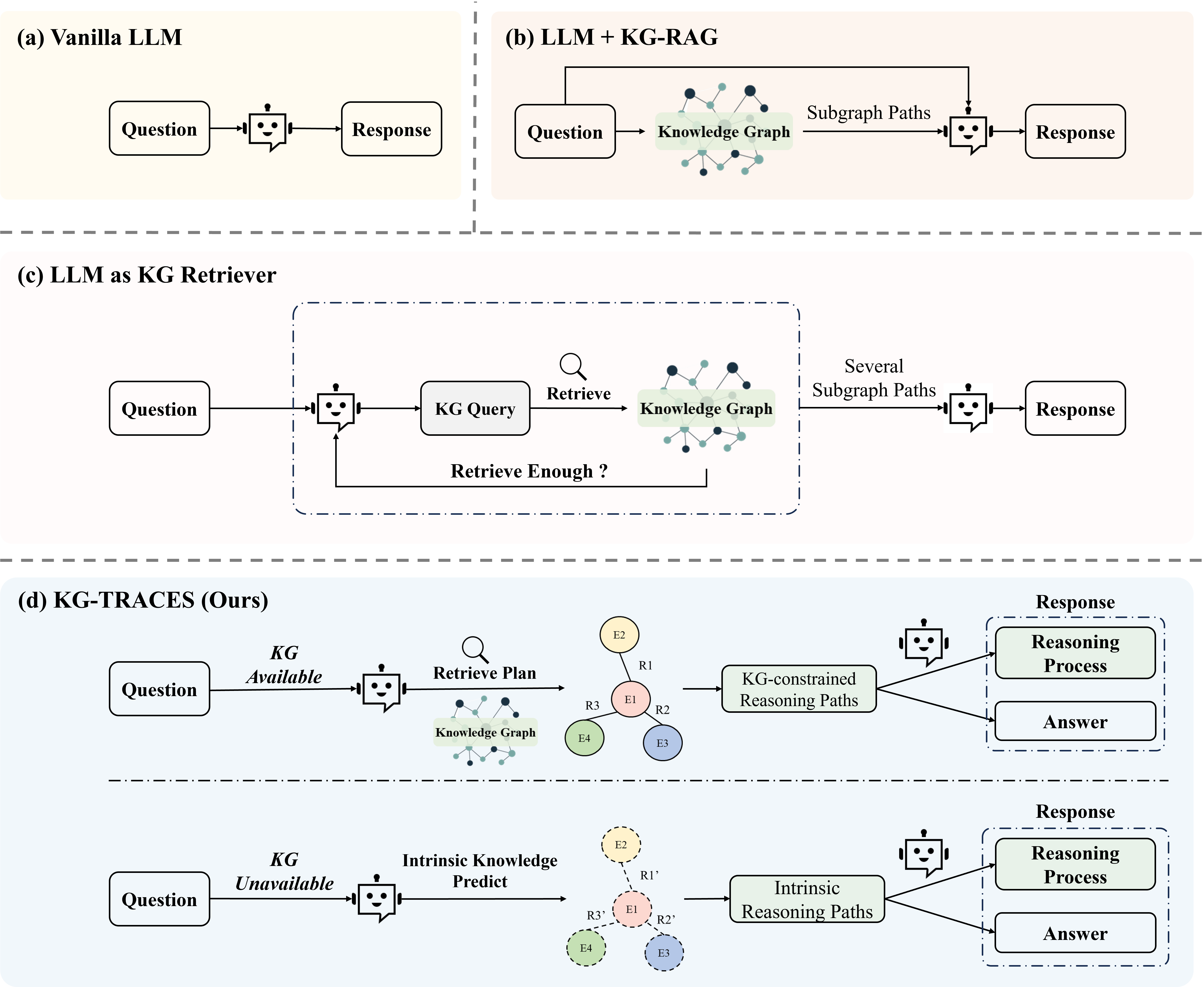}
    \caption{Comparison of representative reasoning methods in LLMs-based frameworks: \textbf{\textit{(a) Vanilla LLMs}}, where the model generates responses directly from the question; \textbf{\textit{(b) LLMs + KG-RAG}}, which uses KG to retrieve relevant subgraph paths to aid the reasoning; \textbf{\textit{(c) LLMs as KG-Retriever}}: where LLMs is a active retriever, querying KG for relevant information, determining whether sufficient knowledge has be retrieved; \textbf{\textit{(d) KG-TRACES (Ours)}}, which can generate faithful and attributable response based on symbolic subgraph reasoning paths under different KG access conditions.}
    \label{fig:intro}
\end{figure}

In this work, we propose \textbf{K}nowledge \textbf{G}raph-constrained \textbf{T}rajectory \textbf{R}easoning \textbf{A}ttribution and \textbf{C}hain \textbf{E}xplanation \textbf{S}upervision (\textbf{KG-TRACES}), a novel framework for supervised reasoning that enables LLMs to generate explainable, attribution-aware reasoning processes under KG access and KG limited scenarios. KG-TRACES is built on the hypothesis that complex reasoning should follow structured and explainable patterns, and LLMs can be trained to internalize such patterns through fine-grained supervision. Therefore, we incorporate reasoning paths derived from KG, which store factual knowledge in the form of triples ⟨subject, relation, object⟩. A reasoning path consists of a sequence of linked triples that connect a question to its answer through intermediate entities and relations. Compared to entity-level facts, relation sequences offer greater abstraction and stability, making them more robust to changes in entity coverage. KG-TRACES is built on three key design principles: (1) supervising LLMs to predict symbolic relation paths and triple-level symbolic reasoning paths that connect questions to answers; (2) training the model to generate step-by-step reasoning processes that are explicitly attributed to either symbolic reasoning paths or just inference based on intrinsic knowledge; and (3) keep faithful reasoning under varying knowledge access conditions—including the presence or absence of KG.

We validate KG-TRACES through extensive experiments on complex reasoning benchmarks, demonstrating significant improvements over previous methods. Furthermore, we evaluate the transferability of our approach by adapting it to specialized domains (medicine) question answering task, where robust and explainable reasoning is critical. To gain deeper insights into the reasoning behaviors enabled by KG-TRACES, we conduct reasoning process visualization analysis~\cite{zhou2025landscape}, 
which describes how the model's reasoning states evolve across stages.

The contributions of this paper can be summarized as follows:
\begin{itemize}[nosep,left=0em]
    \item KG-TRACES is proposed as a unified supervision framework that enables LLMs to perform attribution-aware symbolic reasoning with or without access to external KG.
    \item The framework incorporates a multi-task fine-tuning strategy over constructed structured relation paths, triple paths, and attributable reasoning processes datasets, fostering explainable and attributable reasoning behavior.
    \item Extensive experiments on both general-domain and medical-domain benchmarks validate the effectiveness of KG-TRACES, demonstrating notable reasoning performance improvements due to our supervision design.
\end{itemize}

\section{Related Works}
\label{appendix: related_work}

\subsection{LLM Complex Reasoning with Prompt}

Large language models (LLMs) have demonstrated strong emergent abilities in reasoning, which has led to a surge of interest in prompting-based methods that aim to elicit more clear reasoning processes. CoT prompting encourages step-by-step problem solving by introducing intermediate reasoning steps through examples~\cite{li2025structured}. Building on this idea, Tree-of-Thought (ToT) methods introduce branching and self-evaluation mechanisms, enabling models to explore multiple candidate paths and choose among them~\cite{yao2023tree}. ReAct interleaves reasoning and acting, allowing the model to decide when to retrieve, reflect, or infer~\cite{yao2023react}. Self-consistency further enhances answer robustness by aggregating multiple sampled reasoning chains~\cite{wang2023selfconsistency}.

Although effective, these methods operate entirely through inference-time prompting. The intermediate reasoning steps are unconstrained, lack ground truth alignment, and provide no attribution to external or internal sources. Several works enhance prompting with plan-and-verify strategies~\cite{madaan2023self, zhou2023leasttomost, zelikman2022star}, but still leave the model's reasoning process unsupervised during training. As a result, hallucinations and inconsistencies persist. In contrast, our work introduces training-time supervision over both symbolic paths and natural language reasoning process sequences, enabling models to generate interpretable and attribution-aware outputs.

\subsection{Knowledge Graph Enhanced LLMs Reasoning}

knowledge graph (KG) offer a structured and factual foundation for augmenting the reasoning capabilities of large language models (LLMs). Approaches utilize KG by retrieving and injecting triples directly into model inputs~\cite{zhang2025learning, liu-etal-2024-knowledge-graph}, or by translating questions into KG queries for entity linking and context construction~\cite{lan-jiang-2020-query, ye-etal-2022-rng}. These methods are generally loosely coupled, treating KG as external information sources. More recent work has explored tighter KG–LLM integration. Think-on-Graph (ToG)~\cite{sun2024thinkongraph} introduces an iterative mechanism in which models traverse KG paths step by step via beam search , while Think-on-Graph 2.0 extends this paradigm by combining KG traversal with context retrieval in a hybrid reasoning loop~\cite{ma2024think}. Reasoning on Graphs (RoG)~\cite{luo2024reasoning} introduces a stronger coupling between path planning and reasoning, using retrieved KG paths to supervise model outputs via posterior distillation. Other efforts, such as KELP, focus on scoring and selecting semantically relevant paths to guide model generation~\cite{liu-etal-2024-knowledge-graph}.

However, these approaches generally assume availability of high-coverage KG, and most do not support reasoning under varying access conditions. Moreover, while RoG incorporates path supervision, it does not generate full reasoning processes, nor does it explicitly model provenance or attribution. Our framework, KG-TRACES, differs in three key ways: (1) it trains LLMs to generate both symbolic reasoning paths and natural language processes with step-level attribution; (2) it enabling more robust generalization under KG-present or KG-absent scenarios; and (3) it applies a unified generation scheme regardless of the source of retrieved or predicted paths. These enable improved robustness, interpretability, and domain transferability.

\section{KG-TRACES: Structured Reasoning with KG-Guided Supervision}
\label{KG-TRACES}
KG-TRACES is a structured reasoning framework designed to enhance the explainable and attributable abilities of LLMs via explicit supervision over symbolic reasoning paths and natural language reasoning processes. It operates under a unified generation paradigm applicable in both KG-accessible and KG-absent scenarios.

\subsection{Overview of KG-TRACES Framework}

\begin{figure}[tbh]
    \centering
    \includegraphics[width=1\linewidth]{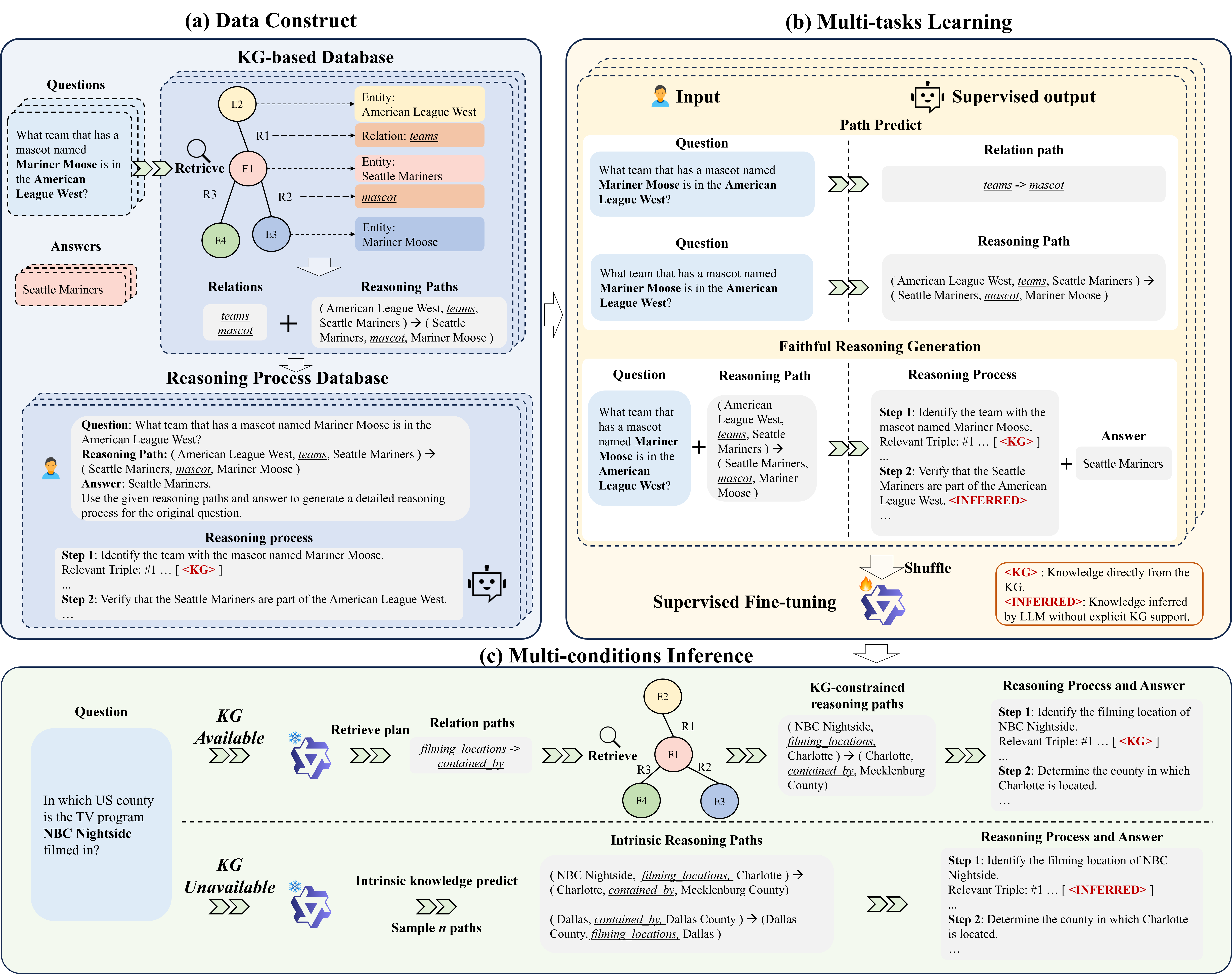}
    \caption{Overview of the KG-TRACES framework. The framework consists of three key components: \textbf{\textit{(a) Data Construction}}: KG-TRACES integrates original datasets with QA data and symbolic reasoning paths from KG to build the KG-based database and reasoning process database for multi-task learning; \textbf{\textit{(b) Multi-task Learning}}: Model is trained for two types of tasks: path prediction (including relation paths prediction used for KG retrieval, and full reasoning paths prediction) and faithful reasoning process generation (supervised model produces attributable, interpretable reasoning processes based on symbolic reasoning paths); \textbf{\textit{(c) Multi-conditions Inference}}: The inference process varies depending on KG availability. With KG access, KG-TRACES predicts relation paths and retrieves whole reasoning paths from  KG, generating faithful reasoning process and answer. Without KG access, KG-TRACES predicts reasoning paths relying on intrinsic knowledge to support faithful reasoning process and answer.}
    \label{fig:framework}
\end{figure}

As Figure~\ref{fig:framework} shows, we supervise LLMs with two type reasoning targets during training: (1) symbolic paths (relation level paths over KG and triple level paths aligned with KG facts), and (2) natural language reasoning processes annotated with step-level attribution provenance. During inference, KG-TRACES can retrieve supporting paths from an external KG when available, or rely on intrinsic knowledge predicted reasoning paths when external KG is unavailable. In both cases, KG-TRACES generates a structured reasoning process and attributes intermediate steps to it's knowledge source.

\subsection{Symbolic Reasoning Paths Prediction Supervision}
\label{sec:symbolic reasoning}

To enable structured and faithful multi-step reasoning, KG-TRACES supervises the model to learn retrieve planning over KG and full reasoning paths which is helpful to answer the question. We frame the retrieve planning as the prediction of symbolic relation paths which link the intermediate relations of reasoning paths from question to its answers.

\vspace{-0.5em}
\paragraph{Notation.}
Let $q$ denote an input question and $\mathcal{G}$ the underlying knowledge graph, composed of factual triples $(e, r, e^{'})$ where $e$ and $e^{'}$ are entities, and $r \in \mathcal{R}$ is a relation. A symbolic relation path\footnote{A symbolic relation path example:  \underline{\textit{teams}} $\rightarrow$ \underline{\textit{mascot}} in Figure~\ref{fig:framework}.} is defined as $r = (r_1, \dots, r_k)$ and a symbolic triple path (reasoning path)\footnote{A symbolic triple path (reasoning path) example: (American League West, \underline{\textit{teams}}, Seattle Mariners) $\rightarrow$ (Seattle Mariners, \underline{\textit{mascot}}, Mariner Moose) in Figure~\ref{fig:framework}.} as $p = ((e_{1}, r_{1}, e_{2}), (e_{2}, r_{2}, e_{3}), \dots, (e_{k}, r_{k}, e_{k+1}))$, both representing multi-hop reasoning trajectories. For each $q$, we denote $\mathcal{R}(q)$ and $\mathcal{T}(q)$ as the retrieved relation and triple path sets, respectively.

Given a dataset $\mathcal{D} = \{(q, \mathcal{R}(q), \mathcal{T}(q))\}$, we train the model to approximate posterior distributions over $r$ and $t$ by minimizing the KL divergence between predicted distributions $P_\theta(\cdot \mid q)$ and empirical distributions $Q(\cdot)$ constructed from retrieval.
The reasoning paths prediction supervision objective consists of two components:

\vspace{-0.5em}
\paragraph{1) Relation Path Distribution Learning.}
The model learns to approximate the posterior distribution $Q(r)$ over retrieved relation paths via KL minimization:

\begin{equation}
\mathcal{L}_{\text{relation}} = \mathbb{E}_{r \sim Q(r)} \left[ \log Q(r) - \log P_\theta(r \mid q) \right]
= \mathrm{D_{KL}}(Q(r) \| P_\theta(r \mid q))
\end{equation}

\vspace{-0.5em}
\paragraph{2) Triple Path Distribution Learning.}
The model similarly matches the posterior distribution $\mathcal{T}(q)$ over triple paths $p$ from retrieval:

\vspace{-1.2em}
\begin{equation}
\mathcal{L}_{\text{triple}} = \mathrm{D_{KL}}(Q(p) \| P_\theta(p \mid q))
\end{equation}
\vspace{-1.2em}

We encourages the model to internalize plausible multi-hop reasoning paths that are discoverable via KG traversal. These symbolic paths serve as structured latent reasoning plan that guide the model toward explainable multi-hop reasoning. Rather than enforcing them as hard-labeled ground truth (which may be incomplete or noisy), we treat these paths as soft supervision signals—posterior samples that regularize the model's prediction behavior. This approach enables KG-TRACES to align its internal reasoning with plausible symbolic paths without being overly constrained by KG completeness, thus remaining robust with partial or sparse KG coverage. When inference, we use beam search to sample several relation paths or triple paths as the paths distribution of model-inferred.

\subsection{Attribution-Aware Reasoning Supervision}
\label{sec:reasoning-process-supervision}

While symbolic reasoning paths reflect model's abstract structured latent plan, practical reasoning requires transforming these plans into concrete, explainable natural language justifications. To this end, KG-TRACES supervises the generation of full reasoning processes annotated with attribution that reflect attribution provenance and causal relevance.

\vspace{-0.5em}
\paragraph{Notation.}
Let $y = (y_1, ..., y_T)$ be a reasoning process sequence. The model generates $y$ conditioned on both the question $q$ and symbolic reasoning paths $p$. $y$ includes attribution labels such as: \texttt{<KG>} and \texttt{<INFERRED>} denoting whether a reasoning step is grounded in retrieved KG knowledge or inferred by model's intrinsic knowledge; \texttt{<EFFECTIVE>} and \texttt{<INEFFECTIVE>} denoting whether a reasoning step is essential for deriving the final answer. These labels help model learn to attribute causal relevance to different parts of the reasoning chain\footnote{Intuitively, not all retrieved or generated paths are useful for answering the question. By marking whether each reasoning step is causally dependent on a specific path, the model is encouraged to distinguish between distractive information and core inferential paths. This token-level attribution allows finer-grained supervision of reasoning quality and helps promote verifiable and goal-directed explanation generation.}. To construct the reasoning process augmented dataset, we employ a general LLM (Qwen-72B-Instruct) to generate explicit step-by-step reasoning processes for each question in the dataset, conditioned on the question, the gold answer, and potential useful reasoning paths. During generation, we instruct the LLM to attribute the basis of each reasoning step using attribution labels (e.g., \texttt{<KG>}, \texttt{<INFERRED>}), enabling fine-grained attribution labels for subsequent training. Detail prompt is in the appendix~\ref{appendix: augment-prompt}.

\vspace{-0.5em}
\paragraph{Objective.}
Conditioned on the question $q$ and a reasoning path $p$, the model is supervised to generate a structured natural language reasoning process $y$ as formula (\ref{L_process}).

\begin{equation}
\label{L_process}
\mathcal{L}_{\text{process}} = - \sum_{j=1}^{T} \log P_\theta(y_j \mid q, p, y_1,\cdots, y_{j-1})
\end{equation}

This enables the model to generate reasoning explanations aligned with symbolic reasoning paths while explicitly learning to attribute evidence source and causal effectiveness.

\vspace{-0.5em}
\paragraph{Unified Training Objective.}
The final objective is the sum of all three components:

\begin{equation}
\mathcal{L}_{\text{KG-TRACES}} = \mathcal{L}_{\text{relation}} + \mathcal{L}_{\text{triple}} + \mathcal{L}_{\text{process}}
\end{equation}

Although all supervision is explicit, the reasoning paths can be viewed as approximations to latent reasoning trajectories that the model would ideally recover on its own. Rather than treating these structures as fixed ground-truths, we treat them as soft supervision to guide the model’s internal reasoning alignment. Special formally, the likelihood of generating a valid reasoning process can be bounded by conditioning on reasoning paths is:

\begin{equation}
    \log P_\theta(y \mid q) \gtrsim \mathbb{E}_{p \sim Q(\cdot)} \left[ \log P_\theta(y_j \mid q, p, y_1,\cdots, y_{j-1} ) \right]
\end{equation}

This training view motivates our use of reasoning paths as soft supervision rather than fixed constraints. Such a soft supervision encourages the model to internalize generalizable reasoning patterns, making it capable of predicting coherent reasoning paths for previously unseen questions. By aligning the model’s internal reasoning with plausible symbolic paths, we approximate latent inference behavior without relying on variational modeling. Unlike prior work that focuses solely on symbolic retrieval planning, we jointly supervise both structured path prediction and attribution-aware reasoning process generation, enabling KG-TRACES to learn faithful and attributable reasoning under a unified objective.

\section{Experiments}
\label{experiment}

\subsection{Experimental Setups}
\vspace{-0.5em}
\paragraph{Tasks and Datasets.}
We conduct comprehensive experiments to evaluate the effectiveness and generalization ability of \textsc{KG-TRACES} across both general-domain and domain-specialized (medicine) reasoning tasks. We evaluate the general reasoning ability of KG-TRACES on two open-domain multi-hop KGQA benchmarks: WebQuestionsSP (WebQSP)~\cite{yih2016value} and Complex WebQuestions (CWQ)~\cite{talmor-berant-2018-web}. To evaluate cross-domain generalization, we additionally use GenMedGPT-5k~\cite{li2023chatdoctor}, a medical QA benchmark constructed from ChatGPT-patient interactions and supported by a curated medical KG of symptoms, diseases, drugs, and treatments~\cite{wen-etal-2024-mindmap}. To validate the quality of our generated data without an explicit filtering stage, we conducted a post-hoc evaluation using a panel of powerful LLMs (GPT-4o, Claude-4-sonnet, and Gemini-2.5-Pro). Across 2,000 samples, our dataset achieved high average scores of 8.1/10 on WebQSP and 7.4/10 on CWQ, confirming its high fidelity. This high quality was further corroborated by a rigorous manual verification on 150 samples (50 from WebQSP, 100 from CWQ), where human evaluation confirmed that 86\% and 83\% of the generated reasoning processes, respectively, were logically sound and factually correct. The details of the datasets are provided in Appendix~\ref{appendix: dataset}.

\vspace{-0.5em}
\paragraph{Implementations.}
We use Qwen2.5-Chat-7B~\cite{yang2024qwen2} as the LLM backbone, which is SFT with multi-tasks on the WebQSP and CWQ for 3 epochs. During inference stage, we sample the top-3 relation paths and top-3 triple paths using beam search for each question. The detailed settings are described in Appendix~\ref{appendix: implementation}.

\vspace{-0.5em}
\paragraph{Baselines.}
In the general domain KGQA benchmarks, we compare KG-TRACES with 25 baselines grouping into 6 categories: 1) Embedding-based methods, 2) Retrieval-augmented methods, 3) Semantic parsing methods, 4) Vanilla LLMs, 5) Prompt Augment LLMs, and 6) LLMs+KG methods. In the medical-domain benchmarks, we compare KG-TRACES with baselines with 3 categories: 1) Retrieval-augmented methods, 2) Vanilla LLMs, 3) LLMs+KG. The details of each baseline are described in Appendix~\ref{appendix: general baselines}.

\vspace{-0.5em}
\paragraph{Evaluation Metrics.}
In the general-domain KGQA benchmarks, we adopt Hits@1 and F1 as our primary evaluation metrics to measure model's performance. In the medical-domain benchmarks, since GenMedGPT-5k consists of generated dialogue between patients and GPT-3.5, conventional string-matching metrics are inadequate. Instead, we adopt LLM-based scoring to assess response quality. The details of medical evaluation metrics are described in Appendix~\ref{appendix: medical metric}.

\subsection{Performance on General Reasoning Tasks}
\begin{figure}[tbh]
    \centering
    \includegraphics[width=1\linewidth]{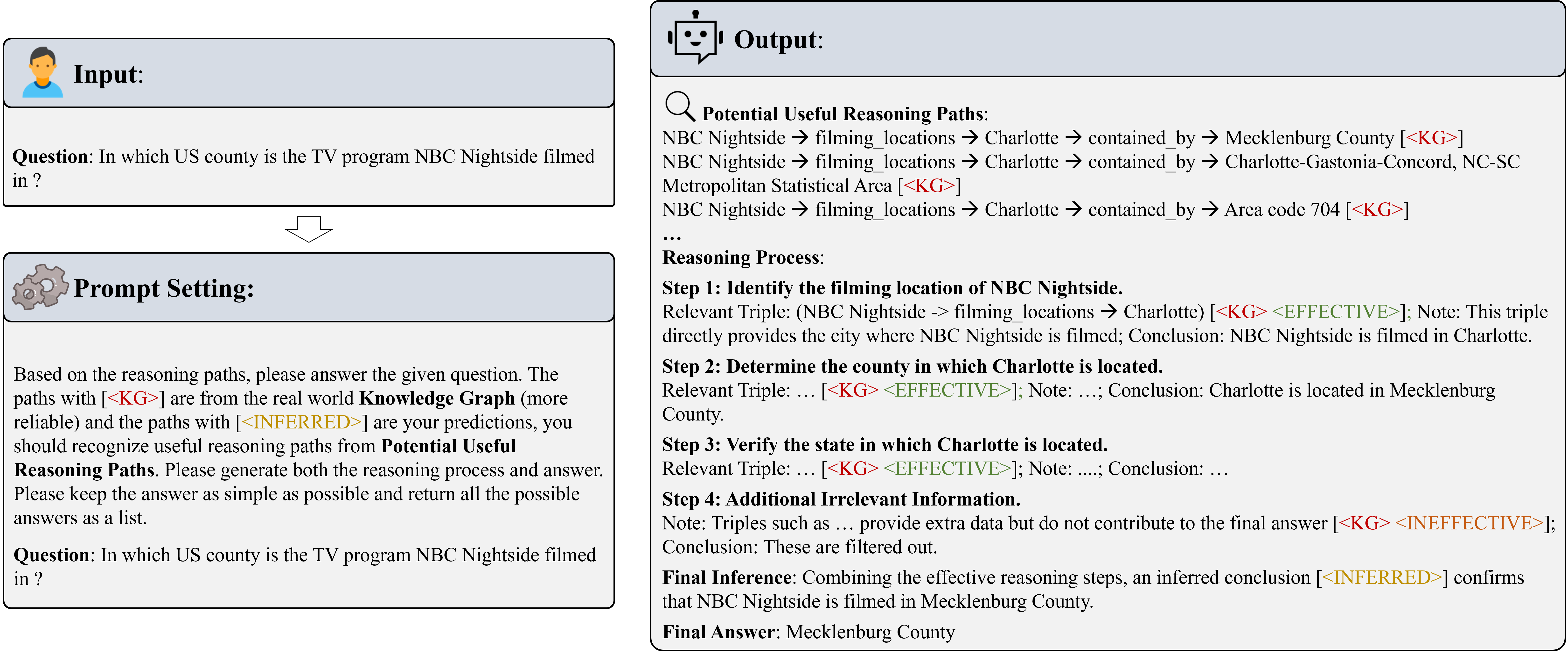}
    \caption{Example of attributable and explainable reasoning of KG-TRACES.}
    \label{fig: QA-text-case}
\end{figure}

As summarized in Table~\ref{tab:baseline-comparison}, our model achieves state-of-the-art performance across both datasets. Compared to the SOTA method RoG~\cite{luo2024reasoning}, which also integrates KG-based planning, KG-TRACES improves Hits@1 by 1.6\% and F1 by 4.7\% on WebQSP, and achieves 4.8\% and 2.1\% gains respectively on CWQ. These gains are more notable on CWQ, which contains a larger proportion of complex 3-hop questions. 
This highlights the value of explicit reasoning process supervision in guiding multi-step reasoning task. Explainable explicit reasoning process supervision may lead to better complex reasoning performance.
And we illustrate a case in Figure~\ref{fig: QA-text-case} to show explainable and attributable reasoning ability of KG-TRACES.

KG-TRACES also consistently surpasses retrieval-based and prompting-based LLM baselines, underscoring the value of explicit supervision over latent reasoning trajectories. In contrast to prior methods that rely solely on prompt heuristics or sparse external knowledge, KG-TRACES enables LLMs to learn explainable and attributable reasoning process aligned with symbolic structure knowledge. These improvements generalize across both simple (WebQSP) and complex (CWQ) reasoning questions, confirming the generalization of our approach.

\begin{table}[t]
\centering
\scriptsize
\caption{Performance comparison on WebQSP and CWQ. Best results in each column are in bold.}
\label{tab:baseline-comparison}
\begin{tabular}{>{\centering\arraybackslash}m{2cm}|>{\raggedright\arraybackslash}p{3.2cm}|>{\centering\arraybackslash}m{1.52cm}>{\centering\arraybackslash}m{1.52cm}|>{\centering\arraybackslash}m{1.52cm}>{\centering\arraybackslash}m{1.52cm}}
\toprule
\multirow{2}{*}{\textbf{Type}} & \multirow{2}{*}{\textbf{Methods}} & \multicolumn{2}{c|}{\textbf{WebQSP}} & \multicolumn{2}{c}{\textbf{CWQ}} \\
& & Hits@1 $\uparrow$ & F1 $\uparrow$ & Hits@1 $\uparrow$ & F1 $\uparrow$ \\
\midrule
\multirow{5}{*}{Embedding} 
& KV-Mem~\cite{miller-etal-2016-key} & 46.7 & 34.5 & 18.4 & 15.7 \\
& EmbedKGQA~\cite{saxena-etal-2020-improving} & 66.6 & - & 45.9 & - \\
& NSM~\cite{he2021improving} & 68.7 & 62.8 & 47.6 & 42.4 \\
& TransferNet~\cite{shi-etal-2021-transfernet} & 71.4 & - & 48.6 & - \\
& KGT5~\cite{saxena-etal-2022-sequence} & 56.1 & - & 36.5 & - \\
\midrule
\multirow{4}{*}{Retrieval} 
& GraftNet~\cite{sun-etal-2018-open} & 66.4 & 60.4 & 38.8 & 32.7 \\
& PullNet~\cite{sun-etal-2019-pullnet} & 68.1 & - & 45.9 & - \\
& SR+NSM~\cite{zhang-etal-2022-subgraph} & 68.9 & 64.1 & 50.2 & 47.1 \\
& SR+NSM+E2E~\cite{zhang-etal-2022-subgraph} & 69.5 & 64.1 & 49.3 & 46.3 \\
\midrule
\multirow{4}{*}{Semantic Parsing}
& SPARQL~\cite{Sun_Zhang_Cheng_Qu_2020} & - & - & 31.6 & - \\
& QGG~\cite{lan-jiang-2020-query} & 73.0 & 73.8 & 36.9 & 37.4 \\
& ArcaneQA~\cite{gu-su-2022-arcaneqa} & - & 75.3 & - & - \\
& RnG-KBQA~\cite{ye-etal-2022-rng} & - & 76.2 & - & - \\
\midrule
\multirow{5}{*}{Vanilla LLMs} 
& Flan-T5-xl~\cite{chung2024scaling} & 31.0 & - & 14.7 & - \\
& Alpaca-7B~\cite{taori2023stanford} & 51.8 & - & 27.4 & - \\
& LLaMA3.1-Chat-8B~\cite{grattafiori2024llama} & 63.4 & 24.7 & 36.9 & 14.2 \\
& Qwen2.5-Chat-7B~\cite{yang2024qwen2}  &  45.7 & 29.3 & 20.2 & 16.1 \\
& ChatGPT~\cite{luo2024reasoning} & 66.8 & - & 39.9 & - \\
\midrule
\multirow{3}{*}{\makecell{Prompt Augmented\\LLMs}}
& LLaMA3.1-Chat-8B + COT & 64.6 & 22.9 & 40.6 & 12.3 \\
& Qwen2.5-Chat-7B + COT &  49.1 & 26.6 & 32.1 & 8.6 \\
& ChatGPT + CoT~\cite{luo2024reasoning} & 75.6 & - & 48.9 & - \\
\midrule
\multirow{5}{*}{LLMs + KG}
& KD-CoT~\cite{wang2023knowledge} & 68.6 & 52.5 & 55.7 & - \\
& UniKGQA~\cite{jiang2023unikgqa} & 77.2 & 72.2 & 51.2 & 49.1 \\
& DECAF~\cite{yu2023decaf} & 82.1 & \textbf{78.8} & - & - \\
& RoG~\cite{luo2024reasoning} & 85.7 & 70.8 & 62.6 & 56.2 \\
& \cellcolor{gray!20} \textbf{KG-TRACES} (ours) & \cellcolor{gray!20} \textbf{87.1} & \cellcolor{gray!20} 74.1 & \cellcolor{gray!20} \textbf{65.6} & \cellcolor{gray!20} \textbf{57.4} \\
\bottomrule
\end{tabular}
\end{table}

\subsection{Analysis Under Limited KG Access}
\label{subsec: analysis_limited_kg}

\vspace{-0.5em}
We evaluate four settings to analyze KG-TRACES under varying KG access conditions: (1) no reasoning path usage (\textit{No KG-aug}), (2) using predicted relation paths for KG retrieval (\textit{KG-aug (rel)}), (3) predicting triple paths without KG (\textit{No KG-aug (triple)}), and (4) combining predicted relations and predicted triple paths (\textit{KG-aug (rel + triple)}).

Results in Table~\ref{tab:limited-kg} show that \textit{KG-aug (rel)} yields the best performance, slightly outperforming \textit{KG-aug (rel + triple)}, highlighting the effectiveness of high-quality reasoning paths. Even without KG access, \textit{KG-aug (triple)} still significantly outperforms LLM-only baselines, demonstrating the model's ability to generate meaningful reasoning paths through intrinsic knowledge. We observe varying trends between datasets: on WebQSP, \textit{KG-aug (triple)} shows limited improvement over \textit{No KG-aug}, while CWQ benefits significantly, with relative gains of 5.1\% in Hits@1 and 4.4\% in F1. These results suggest that even imperfect reasoning paths can provide valuable scaffolding in more complex multi-hop reasoning tasks, especially when questions require navigating longer reasoning chains. This supports the view that structured intermediate representations are especially helpful when navigating long or compositional reasoning chains. Overall, these findings demonstrate the strength of KG-TRACES’s design: by equipping the model to leverage symbolic reasoning paths when KG is available—and fallback gracefully when not—it ensures robustness across different reasoning scenarios.

\begin{table}[hbtp]
\centering
\scriptsize
\setlength{\tabcolsep}{5.5pt}
\caption{
Performance of KG-TRACES under varying KG access conditions on WebQSP and CWQ.}
\label{tab:limited-kg}
\begin{tabular}{
    >{\centering\arraybackslash}m{2.2cm}|
    >{\centering\arraybackslash}m{0.3cm} >{\centering\arraybackslash}m{0.5cm}|
    cccc|cccc
}
\toprule
\multirow{2}{*}{\textbf{Method}} 
& \multicolumn{2}{c|}{\textbf{KG Usage}} 
& \multicolumn{4}{c|}{\textbf{WebQSP}} 
& \multicolumn{4}{c}{\textbf{CWQ}} \\
& Rel & Triple
& Hits@1 $\uparrow$  & F1 $\uparrow$  & Precision $\uparrow$  & Recall $\uparrow$  
& Hits@1 $\uparrow$ & F1 $\uparrow$ & Precision $\uparrow$  & Recall $\uparrow$ \\
\midrule
No KG-aug &  &  & 76.6 & 63.6 & 66.3 & 65.4 & 56.5 & 50.4 & 51.0 & 52.6 \\
No KG-aug (triple) &  & \cmark & 76.8 & 61.8 & 65.6 & 64.7 & 59.4 & 52.6 & 55.4 & 55.2 \\
KG-aug (rel) & \cmark &  & \textbf{87.1} & \textbf{74.1} & \textbf{74.8} & \textbf{79.5} & \textbf{65.6} & \textbf{57.4} & \textbf{57.7} & \textbf{61.7} \\
KG-aug (rel + triple) & \cmark & \cmark & 86.2 & 72.1 & 72.5 & 78.2 & 64.9 & 57.0 & 57.3 & 60.8 \\
\bottomrule
\end{tabular}
\end{table}

\subsection{Additional Evaluation of Supervision and Attribution}
\label{sec:additional}

\paragraph{Ablation of training objectives.}
We ablate each objective on WebQSP and CWQ. Variants without $\mathcal{L}_{relation}$ cannot generate relation paths for retrieval, so they are evaluated without paths and compared with the full model under the same setting.

\begin{table}[h]
\centering
\caption{Objective ablation (Hits@1 / F1). *These variants cannot use KG retrieval.}
\label{tab:abl_obj}
\begin{tabular}{llcc}
\toprule
Variant & Inference & WebQSP & CWQ \\
\midrule
Full KG-TRACES & KG retrieval & 87.1 / 74.1 & 65.6 / 57.4 \\
w/o $\mathcal{L}_{process}$ & KG retrieval & 83.9 / 71.7 & 62.3 / 56.0 \\
w/o $\mathcal{L}_{triple}$ & KG retrieval & 85.5 / 70.8 & 64.6 / 53.1 \\
\midrule
Full KG-TRACES & No path & 76.6 / 63.6 & 56.5 / 50.4 \\
w/o $\mathcal{L}_{relation}$ & No path* & 76.2 / 63.4 & 55.6 / 49.5 \\
QA data only & No path* & 73.9 / 60.4 & 53.8 / 48.4 \\
\bottomrule
\end{tabular}
\end{table}

Removing $\mathcal{L}_{process}$ or $\mathcal{L}_{triple}$ lowers performance under KG retrieval. $\mathcal{L}_{relation}$ makes retrieval possible: without it the model reaches 76.2 / 55.6 Hits@1, while the full model reaches 87.1 / 65.6 with retrieval. QA data only gives the lowest results.

\begin{table}[h]
\centering
\caption{Effect of removing attribution tags (Hits@1 / F1).}
\label{tab:abl_tag}
\begin{tabular}{llcc}
\toprule
Inference setting & Tags & WebQSP & CWQ \\
\midrule
KG relation paths & Full & 87.1 / 74.1 & 65.6 / 57.4 \\
 & No tags & 88.1 / 74.3 & 68.1 / 59.4 \\
\midrule
KG relation + predicted triple & Full & 86.2 / 72.1 & 64.9 / 57.0 \\
 & No tags & 84.9 / 70.4 & 65.8 / 57.0 \\
\bottomrule
\end{tabular}
\end{table}

With KG relation paths alone, removing tags gives similar performance, so the performance gains come mainly from the reasoning supervision rather than the tags. The tags make the source of each step explicit. With KG and predicted triple paths together, tags may help separate the two sources.

\paragraph{Controlled baseline comparison.}
We reproduce RoG with Qwen2.5-7B-Instruct using the same data splits, KG and evaluation protocol

\begin{table}[h]
\centering
\caption{Comparison with reproduced RoG (Qwen2.5-7B-Instruct), Hits@1 / F1.}
\label{tab:rog}
\begin{tabular}{llccc}
\toprule
Setting & Method & Teacher traces & WebQSP & CWQ \\
\midrule
No KG aug. & RoG (reproduced) & No & 50.6 / 38.3 & 32.4 / 27.7 \\
 & KG-TRACES & Yes & 76.6 / 63.6 & 56.5 / 50.4 \\
\midrule
KG retrieval & RoG (reproduced) & No & 80.3 / 67.5 & 50.7 / 45.0 \\
 & KG-TRACES w/o $\mathcal{L}_{process}$ & No & 83.9 / 71.7 & 62.3 / 56.0 \\
 & KG-TRACES & Yes & 87.1 / 74.1 & 65.6 / 57.4 \\
\bottomrule
\end{tabular}
\end{table}

With KG retrieval, KG-TRACES improves over reproduced RoG by 6.8 / 6.6 (Hits@1 / F1) on WebQSP and 14.9 / 12.4 on CWQ. Without teacher-generated traces, it still exceeds RoG by 3.6 / 11.6 Hits@1, so the gain is not explained by the backbone or the extra reasoning process data alone.

\subsection{Influence of Reasoning Paths Number}

\begin{figure}[hbtp]
    \centering
    \includegraphics[width=1\linewidth]{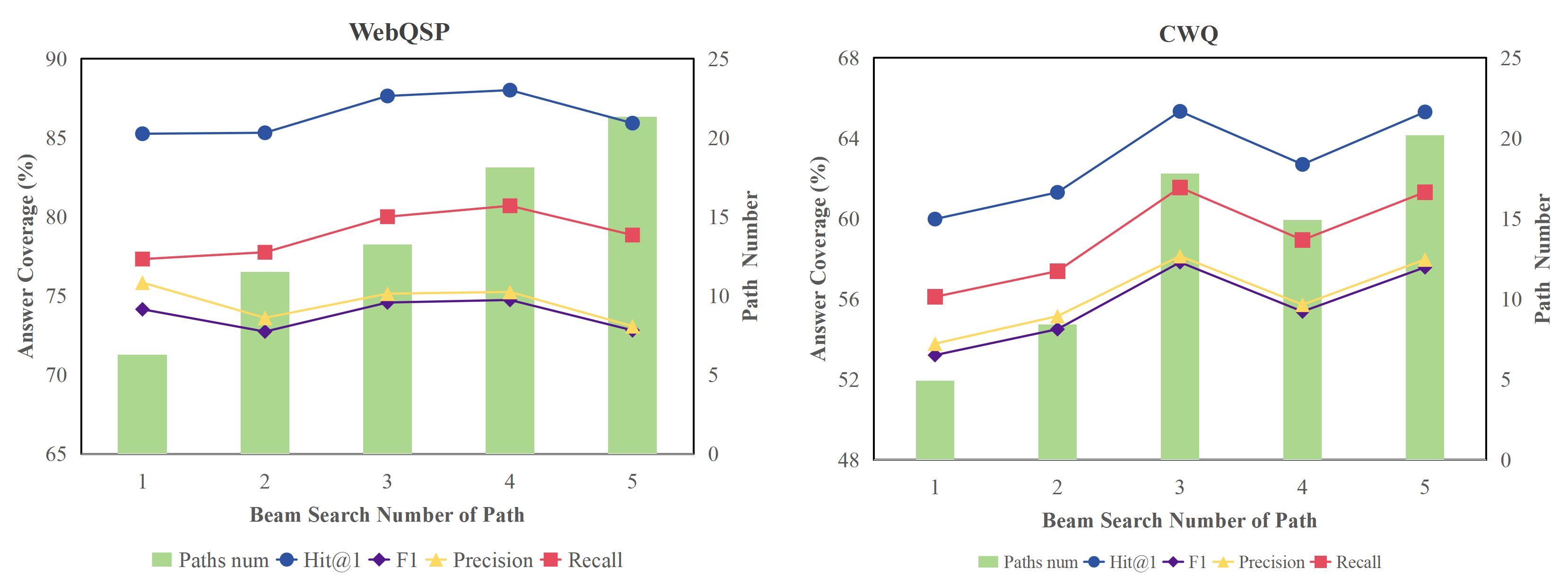}
    \vspace{-1em}
    \caption{KG-TRACES performance comparison based on beam search number of reasoning path. We compare KG-TRACES with relation path using beam search number from 1 to 5 in WebQSP and CWQ.}
    \vspace{-0.75em}
    \label{fig:beam_num-relation_path}
\end{figure}

To understand how the number of candidate reasoning paths affects KG-TRACES, we varied the beam search number from 1 to 5 during relation path prediction. Figure~\ref{fig:beam_num-relation_path} illustrates the performance (Hit@1, F1, Precision, Recall) and the average number of reasoning paths (``Paths num'') on WebQSP and CWQ. As the beam search number increases, more candidate reasoning paths are retrieved, but this does not always translate to better performance.
On WebQSP, key metrics like Hit@1 and F1 generally improve up to a beam search number of 3 or 4, after which performance slightly declines. For the more complex CWQ dataset, these metrics peak at a beam search number of 3, with a noticeable dip at 4 before a partial recovery at 5. These results suggest an optimal beam search size exists (around 3-4 for WebQSP and 3 for CWQ in this setting), balancing sufficient path exploration with the risk of introducing noise from less relevant paths.

\subsection{Cross-Domain Generalization Analysis}

To evaluate the transferability of KG-TRACES to specialized domains, we conduct experiments on dataset \textit{GenMedGPT-5k} without any training. The evaluation details are on the Appendix~\ref{appendix: medical metric} and Appendix~\ref{appendix: medical baselines}.

KG-TRACES Variants. As the model has never observed triples from medical domain, relation path prediction is infeasible. Instead, we only generate full triple paths. Specifically, We add two variants of KG-TRACES: 1) \textit{KG-aug (entity)} (retrieve reasoning paths from question entities), and 2) \textit{KG-aug (entity + triple)} (combine predicted reasoning paths and retrieved reasoning paths).

As shown in Table~\ref{tab:chatdoctor}, KG-TRACES \textit{(No KG-aug)} achieves the best average score (0.7970), surpassing strong baselines such as GPT-4 (0.7596) and MindMap (0.7582) by 4.9\% and 5.1\% respectively. This result underscores KG-TRACES's complex reasoning ability remain effective even in unfamiliar medical domain without any external KG.

Interestingly, we observe that introducing explicit reasoning paths—either predicted paths or retrieved paths both degrades model's performance. We attribute this to domain mismatch and path quality. For the \textit{KG-aug (triple)} variant, KG-TRACES generates paths purely from its internal knowledge, having been trained on general domain KG. These generated paths may contain relation types or compositional patterns that are poorly aligned with medical domain, thereby introducing distributional noise during generation. This mismatch can disrupt the model’s response planning, especially in a zero-shot transfer setting. The resulting 7.5\% average performance drop compared to the \textit{No KG-aug} variant highlights the cost of injecting potentially misleading information. For the \textit{KG-aug (entity)} variant, paths are retrieved based on question entities, such retrieval may suffer from two key limitations: 1) incomplete coverage of the medical KG leads to partial paths, and 2) even when paths exist, they may not lead to the answer. These issues may mislead the model's reasoning.

\begin{table}[t]
\centering
\scriptsize
\setlength{\tabcolsep}{2.9pt}
\caption{
Model performance comparison on medical reasoning task (GenMedGPT-5k). Each score is reported as the mean $\pm$ standard deviation across multiple runs using LLM-based evaluation.
}
\label{tab:chatdoctor}

\resizebox{1\textwidth}{!}{%
\begin{tabular}{l|l|cccccc}
\toprule
\textbf{Type} & \textbf{Method} & \textbf{Relevance} $\uparrow$ & \textbf{Accuracy} $\uparrow$ & \textbf{Completeness} $\uparrow$ & \textbf{Clarity} $\uparrow$ & \textbf{Conciseness} $\uparrow$ & \textbf{Average} $\uparrow$ \\
\midrule
\multirow{3}{*}{Retrieval} 
& BM25 Retriever & $0.70 \pm 0.23$ & $0.67 \pm 0.21$ & $0.56 \pm 0.19$ & $0.85 \pm 0.10$ & $0.75 \pm 0.10 $ & $0.70 \pm 0.16$ \\
& Embedding Retriever & $0.74 \pm 0.22$ & $0.70 \pm 0.18$ & $0.59 \pm 0.18$ & $0.87 \pm 0.09$ & $0.76 \pm 0.09$ & $0.73 \pm 0.15$ \\
& KG Retriever & $0.72 \pm 0.24$ & $0.69 \pm 0.20$ & $0.57 \pm0.20 $ & $0.87 \pm 0.09$ & $0.77 \pm 0.09$ & $0.72 \pm 0.16$ \\
\midrule
\multirow{2}{*}{Vanilla LLMs} 
& ChatGPT & $0.80 \pm 0.14$ & $0.76 \pm 0.10$ & $0.62 \pm 0.12$ & $0.91 \pm 0.05$ & $0.80 \pm 0.05$ & $0.78 \pm 0.09$ \\
& GPT-4 & $0.77 \pm 0.21$ & $0.73 \pm 0.18$ & $0.59 \pm 0.19$ & $0.90 \pm 0.06$ & $\mathbf{0.81} \pm 0.06$ & $0.76 \pm 0.13$ \\
\midrule
\multirow{1}{*}{LLMs + KG} 
& MindMap & $0.77 \pm 0.21$ & $0.72 \pm 0.18$ & $0.61 \pm 0.18$ & $0.89 \pm 0.08$ & $0.80 \pm 0.07$ & $0.76 \pm 0.14$ \\
\midrule
\multirow{4}{*}{ \cellcolor{white}\makecell{KG-TRACES\\ Variants}}
& No KG-aug & $\mathbf{0.83} \pm 0.20$ & $\mathbf{0.77} \pm 0.20$ & $\mathbf{0.68} \pm 0.16$ & $\mathbf{0.92} \pm 0.08$ & $0.79 \pm 0.07$ & $\mathbf{0.80} \pm 0.13$ \\
& No KG-aug (triple) & $0.76 \pm 0.22$ & $0.71 \pm 0.22$ & $0.63 \pm 0.19$ & $0.87 \pm 0.11$ & $0.73 \pm 0.10$ & $0.74 \pm 0.16$ \\
& KG-aug (entity) & $0.77 \pm 0.22$ & $0.70 \pm 0.21$ & $0.63 \pm 0.17$ & $0.88 \pm 0.10$ & $0.74 \pm 0.09$ & $0.75 \pm 0.15$ \\
& KG-aug (entity + triple) & $0.77 \pm 0.21$ & $0.70 \pm 0.21$ & $0.63 \pm 0.17$ & $0.87 \pm 0.10$ & $0.73 \pm 0.11$ & $0.74 \pm 0.15$ \\
\bottomrule
\end{tabular}
}
\end{table}

\subsection{Reasoning Process Quality and Visualization}

We analyze the intermediate reasoning steps of KG-TRACES to understand whether multi-tasks supervision fosters structured and convergent reasoning behaviors. Following the prior work~\cite{zhou2025landscape}, we segment the reasoning process into stages and visualize the model's evolving reasoning states, along with three metrics—\textit{consistency}, \textit{uncertainty}, and \textit{perplexity}—to quantitatively track reasoning progression\footnote{We calculate \textit{consistency} between reasoning process thoughts of each stage and final thoughts, \textit{uncertainty} and \textit{perplexity} between question and reasoning process thoughts of each stage}. Metric definitions and visualization are detailed in Appendix~\ref{appendix: reasoning quality metric}.

\begin{figure}[!t]
    \centering
    \includegraphics[width=1\linewidth]{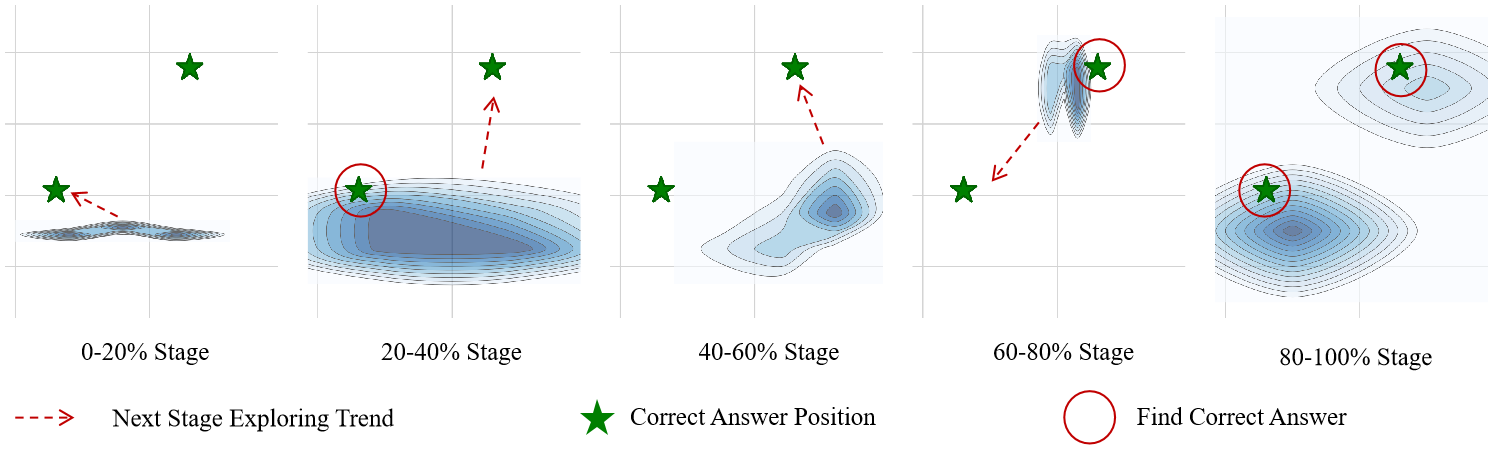}
    \vspace{-5pt}
    \caption{
    Visualization of model reasoning thoughts for a representative case in WebQSP. Darker color denotes higher reasoning process thoughts distribution density of the region. As reasoning progresses, thoughts distributions become sharper and align more closely with answers. \textbf{Example:}(\textit{Question:} what year did the LA kings win the cup? \textit{Answers:} 2012 Stanley Cup Finals, 2014 Stanley Cup Finals.) }
    \label{fig:webqsp-case-reasoning-trajectory}
\end{figure}

\vspace{-0.5em}
\paragraph{Case-level progression analysis.}
Figure~\ref{fig:webqsp-case-reasoning-trajectory} illustrates the trajectory of a representative QA instance over five reasoning stages. We observe that early thoughts are scattered and uncertain, with thoughts exploring in a large latent space. As reasoning progresses, reasoning thoughts distributions become increasingly concentrated around the correct answers, supported by the distribution density increasing in the correct answer regions. Specifically, in the early stages (0-20\% and 20-40\%), the reasoning process exhibits substantial exploration, where the model is uncertain about the answer. In subsequent stages (40-60\%, 60-80\%), the reasoning trajectory begins to narrow as the model identifies more promising directions. The model's exploration in the latent space progressively refining the reasoning process. By the final stages (80-100\%), KG-TRACES reaches all correct answers. This highlights how KG-TRACES, through explicit reasoning supervision and attribution-aware processes, guides the model to a stable and accurate conclusion.

\begin{figure}[!t]
    \centering
    \includegraphics[width=1\linewidth]{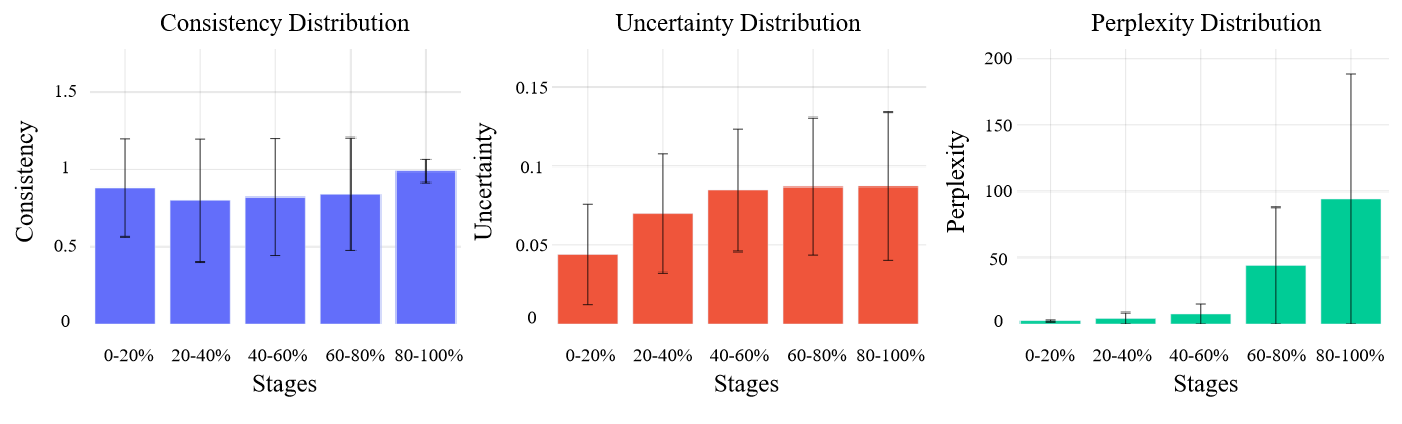}
    \vspace{-5pt}
    \caption{Visualization of step-wise reasoning process metrics distribution of KG-TRACES in WebQSP}
    \label{fig:webqsp-distribution-reasoning-trajectory}
\end{figure}

\begin{figure}[!t]
    \centering
    \includegraphics[width=1\linewidth]{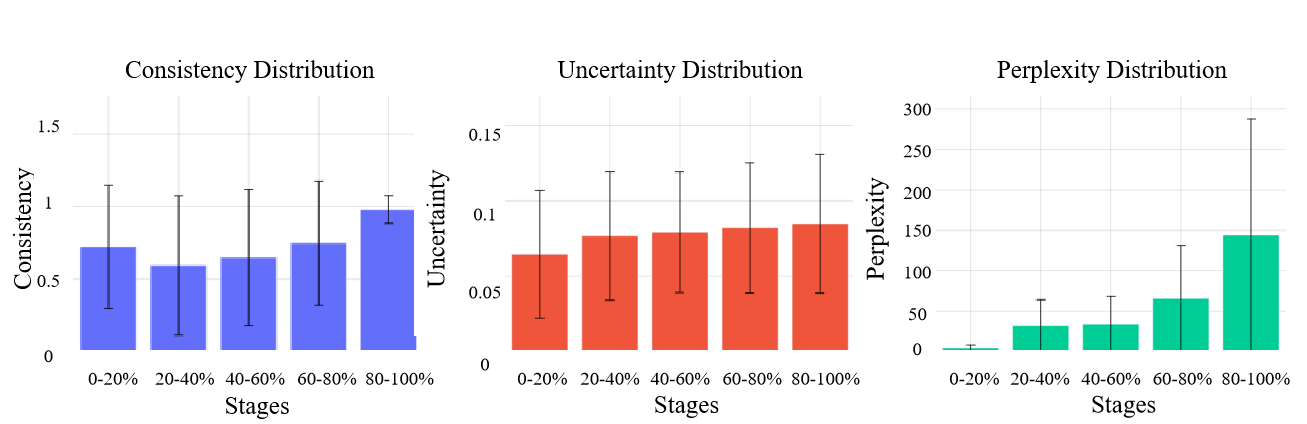}
    \caption{Visualization of step-wise reasoning metrics distributions of KG-TRACES in CWQ}
    \label{fig:cwq-distribution-reasoning-trajectory}
\end{figure}

\vspace{-0.5em}
\paragraph{Stage-wise metric distribution analysis.}
To further quantify this stage-wise behavior, we compute the mean and standard deviation for \textit{Consistency}, \textit{Uncertainty} and \textit{Perplexity} across a subset of examples from the test dataset of WebQSP and CWQ. As shown in the Figure~\ref{fig:webqsp-distribution-reasoning-trajectory} and Figure~\ref{fig:cwq-distribution-reasoning-trajectory}, \textit{consistency} steadily increases across stages, meaning intermediate states become increasingly consistent with the final conclusion. \textit{Uncertainty} and \textit{Perplexity} also steadily increases across stages, indicating that the model explores more latent reasoning space as the reasoning process unfolds. These highlight the effectiveness of KG-TRACES in guiding the model toward explainable and attributable conclusions over reasoning steps. The analysis of more case and result of WebQSP and CWQ will be discussed in Appendix~\ref{appendix: reasoning quality metric}.

\vspace{-1em}

\section{Conclusion}
We present KG-TRACES, a unified framework for training large language models to perform explainable, attributable reasoning guided by reasoning paths from knowledge graph. By supervising models on relation paths, triple paths, and attribution-aware reasoning processes, KG-TRACES enables faithful multi-step inference across both general and domain-specific scenarios. Through extensive experiments, we show that KG-TRACES achieves strong performance under varying KG accessibility, transfers effectively to unseen medical QA tasks, and produces stable reasoning trajectories. Our results highlight the value of structured symbolic path and attribution-aware reasoning processes supervision for enhancing both the accuracy and transparency of language model reasoning.

\section{Limitations and broader impacts}
\label{appendix: limitations}
While KG-TRACES demonstrates strong performance on both general and specialized QA tasks, several limitations remain. First, the reliance on symbolic supervision requires access to high-quality KG-derived paths, which may be incomplete or noisy in low-resource domains. Second, our current evaluation focuses on multi-hop QA task; extending to more diverse reasoning types (e.g., math, procedural) warrants further study. Future work includes exploring semi-supervised or reinforcement learning for symbolic path induction, and scaling our framework to more reasoning scenarios. Overall, this work has the potential to enhance interpretability and robustness in knowledge-intensive applications, facilitating more transparent and trustworthy AI systems.

\small
\bibliographystyle{plain}  
\bibliography{reference.bib}



\newpage
\appendix


\section*{Appendix} 

\startcontents 
\printcontents{}{1}{} 

\newpage

\section{Dataset Details}
\label{appendix: dataset}
\subsection{General Reasoning Dataset}
To evaluate KG-TRACES's generative reasoning performance, we use two KGQA datasets: WebQuestionSP (WebQSP)\cite{yih2016value} and Complex WebQuestions (CWQ)~\cite{talmor-berant-2018-web} as our main benchmark. Both datasets are grounded in the Freebase knowledge graph~\cite{bollacker2008freebase}. We follow previous works~\cite{luo2024reasoning} to use the same train and test splits for fair comparison. The statistic of the datasets are given in Table~\ref{tab:main-dataset-split}. The statistics of the answer numbers and reasoning hops are presented in Table 7 and Table 8, respectively.

To demonstrate the high quality of the reasoning processes generated by Qwen-72B-Instruct, we performed a random sample post-hoc validation. This audit serves to quantify the reliability and coherence of our automatically constructed dataset, justifying its direct use for model training without an intermediate filtering step.

Our validation methodology involved a committee of powerful, closed-source LLMs (GPT-4o, Claude-4-sonnet, and Gemini-2.5-Pro) acting as evaluators. We randomly sampled 500 instances from the WebQSP subset and 1,500 from the CWQ subset. Each instance, comprising the question, ground truth answer, and the generated reasoning process, was scored by the evaluators on a scale of 0-10.

The prompt used for this validation is presented in Table~\ref{tab:score_prompt_resoning_process}. It was designed to be simple and direct, focusing on the core correctness and quality of the reasoning.

\begin{table}[h]
\scriptsize
\centering
\caption{Score prompt for validation of reasoning process quality.}
\label{tab:score_prompt_resoning_process}
\renewcommand{\arraystretch}{1.2}
\begin{tabular}{p{0.97\linewidth}}
\toprule
\textbf{Score Prompt (Reasoning Process Quality)} \\
\midrule
Given the following question, ground truth answer, and a reasoning process, please evaluate the correctness and quality of the reasoning on a scale of 0 to 10. Provide only the score as an integer. 

Question: \{question\} \\

Ground Truth Answer: \{ground truth answer\} \\

Reasoning process: \{llm generated reasoning process\} \\

\bottomrule
\end{tabular}
\end{table}

\begin{table}[htbp]
\scriptsize 
\centering
\caption{Statistics of two generative reasoning datasets.}
\label{tab:main-dataset-split}
\begin{tabular}{
>{\centering\arraybackslash}m{2.35cm}
>{\centering\arraybackslash}m{2.35cm}
>{\centering\arraybackslash}m{2.35cm}
>{\centering\arraybackslash}m{2.35cm}
>{\centering\arraybackslash}m{2.35cm}}
\toprule
\textbf{Dataset} & \textbf{Train} & \textbf{Validation} & \textbf{Test} & \textbf{Max hop} \\
\midrule
WebQSP & 2826 & 246 & 1628 & 2 \\
CWQ    & 27626 & 3519 & 3531 & 4 \\
\bottomrule
\end{tabular}
\end{table}

\begin{table}[htbp]
\scriptsize 
\centering
\caption{Answer count distribution on WebQSP and CWQ. $Ans_{num}$ denotes the number of answers per question.}
\label{tab:answer_count_stats}
\begin{tabular}{
>{\centering\arraybackslash}m{2.35cm}
>{\centering\arraybackslash}m{2.35cm}
>{\centering\arraybackslash}m{2.35cm}
>{\centering\arraybackslash}m{2.35cm}
>{\centering\arraybackslash}m{2.35cm}}
\toprule
\textbf{Dataset} & $Ans_{num} = 1$ & $2 \leq Ans_{num} \leq 4$ & $5 \leq Ans_{num} \leq 9$ & $Ans_{num} \geq 10$ \\
\midrule
WebQSP & 51.2\% & 27.4\% & 8.3\% & 12.1\% \\
CWQ & 70.6\% & 19.4\% & 6.0\% & 4.0\% \\
\bottomrule
\end{tabular}
\end{table}

\begin{table}[H]
\scriptsize 
\centering
\caption{QA hops Statistics of WebQSP and CWQ.}
\label{tab:hop_statistics}
\begin{tabular}{
>{\centering\arraybackslash}m{3.06cm}
>{\centering\arraybackslash}m{3.06cm}
>{\centering\arraybackslash}m{3.06cm}
>{\centering\arraybackslash}m{3.06cm}}
\toprule
\textbf{Dataset} & \textbf{1 hop} & \textbf{2 hop} & $\boldsymbol{\geq}$\textbf{3 hop} \\
\midrule
WebQSP\_{train} & 62.89\% & 37.11\% & 0\%  \\
WebQSP\_{test} & 64.11\% & 35.88\% & 0\% \\
CWQ\_{train} & 24.65\% & 57.23\% & 18.12\% \\
CWQ\_{test} & 22.11\% & 57.89\% & 20.00\% \\
\bottomrule
\end{tabular}
\end{table}

\subsection{Medical Domain Dataset}
GenMedGPT-5k is a synthetic dataset consisting of 5,000 dialogues between patients and GPT-3.5, generated based on a structured disease knowledge base~\cite{li2023chatdoctor}\footnote{https://github.com/Kent0n-Li/ChatDoctor}. Each dialogue begins with a patient question derived from real-world medical consultations in the iCliniq database, describing specific symptoms or health concerns. The model-generated responses include detailed medical reasoning, covering diagnosis, symptoms, treatment recommendations, and suggested medical tests. For fair comparison , we use the same 714 dialogues of previous work~\cite{wen-etal-2024-mindmap} to construct the test dataset used in our medical domain experiments.

\section{Prompt Details}
\label{appendix: prompt_details}

\subsection{Multi-tasks Supervision Dataset Construct Prompts}
\label{appendix: augment-prompt}

To generate training data for multi-tasks supervision, we design structured prompts that guide LLMs to produce symbolic relation paths, triple paths and explainable intermediate reasoning processes. These prompts are used in three contexts:

\begin{itemize}[nosep, left=0em]
    \item \textbf{Full reasoning process generation}: The model is asked to simulate multi-step reasoning given the question, answer and potential supporting KG paths, while labeling each step's supporting attribution (KG or model-inferred) and utility.
    \item \textbf{Relation path construction}: The model generates a valid sequence of relations that connects question entities to the answer.
    \item \textbf{Triple path construction}: The model outputs explicit triple-level paths, capturing symbolic reasoning paths supervision.
\end{itemize}

We show the detail prompts used for each setting in Table~\ref{tab:resoning_process_constrcut_prompt}, ~\ref{tab:relation_path_construct_prompt} and ~\ref{tab:triple_path_construct_prompt}.

\begin{table}[htbp]
\scriptsize 
\centering
\caption{Prompt template for constructing structured reasoning processes dataset of KG-TRACES}
\label{tab:resoning_process_constrcut_prompt}
\begin{tabular}{p{0.97\linewidth}}
\toprule
\textbf{KG-TRACES Reasoning Process Construction Prompt} \\
\midrule
\#\#\# Question: \\
\{question\} \\

\vspace{0.5em}
\#\#\# Answer: \\
\{answer\} \\

\vspace{0.5em}
\#\#\# Potential useful reasoning path: \\
The following reasoning paths are provided to help you understand relationships among entities and derive an answer: \\
\{reasoning\_paths\} \\

\vspace{0.5em}
\#\#\# Task Instructions: \\
1. Goal:
\begin{itemize}
  \item Use the given reasoning paths and answer to generate a detailed reasoning process for the original question, explicitly indicating the source of knowledge (e.g., from KG or inferred by LLMs).
  \item Enhance the reasoning process by including special tokens to label each path's source and effectiveness:
  \begin{itemize}
    \item <KG>: Knowledge directly from the knowledge graph.
    \item <INFERRED>: Knowledge inferred by LLMs without explicit KG support.
    \item <EFFECTIVE> / <INEFFECTIVE>: Whether the path effectively contributes to the final answer.
  \end{itemize}
\end{itemize}

2. Specific Requirements:
\begin{itemize}
  \item Path Selection and Labeling:
  \begin{itemize}
    \item Filter out unnecessary paths: Only select paths directly relevant to the question.
    \item Ignore paths marked as <INEFFECTIVE> when getting the final answer.
    \item Label each selected path using the special tokens.
  \end{itemize}
  \item Dynamic Knowledge Utilization:
  \begin{itemize}
    \item If no KG path applies, allow LLMs to infer logical connections using <INFERRED>, clearly marked.
  \end{itemize}
\end{itemize}

3. Output Format: \\
**Reasoning Process**: [Output reasoning process here] \\

\#\#\# Example: \\

[Input] \\
**Question**: Which film directed by Christopher Nolan starred Leonardo DiCaprio and was released in 2010? \\
**Answer**: Inception \\
**Retrieved Triples**: \\
1. Leonardo DiCaprio → film.actor.film → m.12345 \\
2. m.12345 → film.director → Christopher Nolan \\
... \\
19. m.00000 → film.release\_date → 2017 \\

[Output] \\
**Reasoning Process**: \\
Step 1: Identify film starring Leonardo DiCaprio. \\
- Relevant Triple: \#1 [<KG> <EFFECTIVE>] \\
- Note: Triples \#10/\#13 [<KG> <INEFFECTIVE>] \\
Step 2: Directed by Christopher Nolan: \#2 [<KG> <EFFECTIVE>] \\
... \\
Final Answer: Inception \\

\bottomrule
\end{tabular}
\end{table}

\begin{table}[htbp]
\scriptsize 
\centering
\caption{Prompt template for constructing symbolic relation paths dataset of KG-TRACES}

\label{tab:relation_path_construct_prompt}
\renewcommand{\arraystretch}{1.2}
\begin{tabular}{p{0.97\linewidth}}
\toprule
\textbf{Path Construction Prompt (Relation Path)} \\
\midrule
Please generate a valid reasoning relation path that can be helpful for answering the following question. \\
**Question**: \{question\} \\

\bottomrule
\end{tabular}
\end{table}

\begin{table}[htbp]
\scriptsize 
\centering
\caption{Prompt template for constructing symbolic triple paths dataset of KG-TRACES}
\label{tab:triple_path_construct_prompt}
\renewcommand{\arraystretch}{1.2}
\begin{tabular}{p{0.97\linewidth}}
\toprule
\textbf{Path Construction Prompt (Triple Path)} \\
\midrule
Please generate a valid reasoning triple path that can be helpful for answering the following question. \\
**Question**: \{question\} \\

\bottomrule
\end{tabular}
\end{table}

\subsection{Inference Prompts}
\label{appendix:inference-prompts}

During evaluation, we apply two types of prompts depending on the model's access to external knowledge:

\begin{itemize}[nosep, left=0em]
    \item \textbf{No-path prompt}: The model answers each question independently, without any external symbolic reasoning paths information.
    \item \textbf{Path-informed prompt}: The model is supplied with retrieved or model generated reasoning paths, and must generate both the reasoning process and final answer while distinguishing between factual and inferred knowledge.
\end{itemize}

These inference prompts, shown in Table~\ref{tab:inference-no_path} and~\ref{tab:inference_reasoning_path}, are used consistently in all experimental evaluations of WebQSP and CWQ.

\begin{table}[htbp]
\scriptsize 
\centering
\caption{Inference prompt for direct answer generation without any reasoning paths.}
\label{tab:inference-no_path}
\renewcommand{\arraystretch}{1.2}
\begin{tabular}{p{0.97\linewidth}}
\toprule
\textbf{Inference Prompt (No Reasoning Path)} \\
\midrule
Please answer the following questions. Please keep the answer as simple as possible and return all the possible answers as a list. \\
**Question**: \{question\} \\

\bottomrule
\end{tabular}
\end{table}

\begin{table}[htbp]
\scriptsize 
\centering
\caption{Inference prompt for reasoning process and answer generation with reasoning paths.}
\label{tab:inference_reasoning_path}
\begin{tabular}{p{0.97\linewidth}}
\toprule
\textbf{Inference Prompt (With Reasoning Paths)} \\
\midrule
Based on the reasoning paths, please answer the given question. The paths with [<KG>] are from the real world **Knowledge Graph** (more reliable) and the paths with [<INFERRED>] are your predictions. You should recognize useful reasoning paths from **Potential Useful Reasoning Paths**. Please generate both the reasoning process and answer. Please keep the answer as simple as possible and return all the possible answers as a list. \\

**Potential Useful Reasoning Paths**: \{reasoning\_paths\} \\
**Question**: \{question\} \\

\bottomrule
\end{tabular}
\end{table}

\section{Training and Implementation Details}
\label{appendix: implementation}
For KG-TRACES, we use Qwen2.5-Chat-7B~\cite{yang2024qwen2} as the LLM backbone, which is instruction finetuned on the training split of WebQSP and CWQ with Freebase for 3 epochs. In addition to the original datasets, we augment the training data by incorporating two additional types: QA-based and path-based SFT data (QA-based data consists of question-answer pairs, while path-based SFT data includes reasoning paths derived from KG). This data augmentation helps improve the model's generalization and reasoning capabilities. The optimization objective during fine-tuning is minimizing the loss between generated text and target text. We set the maximum context length to 4096, padding each batch to match the longest sequence in that batch. The batch size is set to 4, the learning rate is set to 2e-5, and the gradient accumulation step is set to 16. We use the cosine learning rate scheduler policy with the warmup ratio set to 0.03. For LoRA fine-tuning, we utilized DeepSpeed, BF16 data type, and gradient checkpointing technology. The training is conducted on 6 A100-80G GPUs for 30 hours.

During inference stage of general reasoning task, we first adopt the model to generate top-K relation paths and triple paths with the highest probability. When inference with the predicted relation paths, we utilized the same method with~\cite{luo2024reasoning} to retrieve reasoning paths in KG for each question to get reasoning paths, and prompt model to response. When inference with the predicted triple paths, we just link the triple paths according to head and tail entity to get reasoning paths.

During inference in the medical task, KG-TRACES is directly evaluated on the GenMedGPT-5k dataset without any further fine-tuning on medical knowledge. Since the model has not been exposed to the medical KG during training, it cannot generate relation paths; thus, only triple paths are predicted. We obtain reasoning paths by linking predicted triples through head–tail entity alignment, similar to the method used in the general reasoning task. For KG-augmented variants, we additionally retrieve reasoning paths by conducting entity linking based on the question entities to extract relevant subgraphs from the medical KG EMCKG follow previous work~\cite{wen-etal-2024-mindmap}. A fixed number (30) of reasoning paths are randomly sampled as retrieved context. All model response are generated conditioned on these reasoning paths via prompting.

\vspace{-1em}
\subsection{General Reasoning Task Baselines}
\label{appendix: general baselines}

For the general reasoning tasks (WebQSP and CWQ), we evaluate KG-TRACES against a range of baseline methods that span several categories follow the previous work~\cite{luo2024reasoning}:

\begin{itemize}[nosep, left=0em]
  \item \textbf{Embedding-based methods:} These methods rely on embedding-based representations to match questions with relevant knowledge. This includes models like KV-Mem~\cite{miller-etal-2016-key}, which stores entities and their relationships in memory for retrieval, and EmbedKGQA~\cite{saxena-etal-2020-improving}, which encodes knowledge graph into embeddings for question answering. Other models in this category include NSM~\cite{he2021improving}, TransferNet~\cite{shi-etal-2021-transfernet} and KGT5~\cite{saxena-etal-2022-sequence}.
  
  \item \textbf{Retrieval-augmented methods:} These methods retrieve relevant information from knowledge graph or external databases to aid in answering questions. Notable models in this category include GraftNet~\cite{sun-etal-2018-open}, PullNet~\cite{sun-etal-2019-pullnet}, and the more recent SR+NSM~\cite{zhang-etal-2022-subgraph} and SR+NSM+E2E~\cite{zhang-etal-2022-subgraph}.
  \item \textbf{Semantic parsing methods:} These methods transform questions into formal queries over knowledge graph. SPARQL-based approaches, such as QGG~\cite{lan-jiang-2020-query}, ArcaneQA~\cite{gu-su-2022-arcaneqa} and RnG-KBQA~\cite{ye-etal-2022-rng}, use graph querying to retrieve answers directly from knowledge graph.
  
  \item \textbf{Vanilla LLMs:} These methods rely solely on LLMs for question answering, including models like ChatGPT, Flan-T5~\cite{chung2024scaling}, and Alpaca-7B~\cite{taori2023stanford}.
  
  \item \textbf{Prompt Augmented LLMs.} To evaluate the impact of prompting strategies on LLMs, we additionally consider a set of \textit{Prompt Augmented LLMs} that incorporate CoT reasoning instructions into the input. Specially, we evaluate models like LLaMA3.1-Chat-8B~\cite{grattafiori2024llama}, Qwen2.5-Chat-7B~\cite{yang2024qwen2}, and ChatGPT with a CoT prompt prepended to the question, encouraging the model to reason step by step before outputting an answer.

  \item \textbf{LLMs+KG:} These methods combine LLMs with knowledge graph to improve reasoning over structured data. Models such as KD-CoT~\cite{wang2023knowledge}, UniKGQA~\cite{jiang2023unikgqa}, and RoG~\cite{luo2024reasoning} fall into this category.
\end{itemize}

For each of the baselines, we report Hits@1 and F1 scores on both WebQSP and CWQ datasets to evaluate their performance.

\subsection{Medical Reasoning Task Baselines}
\label{appendix: medical baselines}

In the medical reasoning domain, we compare KG-TRACES with baselines designed specifically for medical question answering. These baselines include retrieval-based methods, vanilla LLMs, and methods that integrate knowledge graph for medical reasoning. The following categories represent the main baselines evaluated:

\begin{itemize}[nosep, left=0em]
  \item \textbf{Retrieval-augmented methods:} These models enhance question answering by retrieving relevant medical knowledge from external sources. Notable methods in this category include BM25 Retriever and Embedding Retriever~\cite{wen-etal-2024-mindmap}, which retrieve relevant medical knowledge for ChatGPT to generate responses.
  
  \item \textbf{Vanilla LLMs:} This category includes LLMs that do not use external knowledge graph but rely solely on the model’s internal knowledge base. We use models such as ChatGPT and GPT-4 for comparison, as these models have shown strong performance in natural language understanding and generation tasks.
  
  \item \textbf{LLMs + KG:} These models combine LLMs with external medical knowledge graph to improve the quality of medical reasoning. MindMap~\cite{wen-etal-2024-mindmap} integrates symbolic reasoning using KG-based prompts and has been previously demonstrated to perform well in medical question answering tasks.
\end{itemize}

For the medical reasoning task, we evaluate the performance of each model using the same evaluation criteria and metrics, such as Relevance, Accuracy, Completeness, Clarity, and Conciseness, which are described in detail in Appendix~\ref{appendix: medical metric}.

\section{Metrics and Scoring Details}
\label{appendix: metrics}

\subsection{Medical Reasoning Task Metrics}
\label{appendix: medical metric}

All models are scored using the \textit{qwen-plus}\footnote{\textit{qwen-plus} refers to the API-accessible version of Alibaba's language model, evaluated via \url{https://www.aliyun.com/product/bailian}.} API, with each response rated 3 times and averaged for stability. Detail evaluation prompt is in the Table~\ref{tab:medical-eval-prompt}.

\begin{table}[htbp]
\scriptsize 
\centering
\caption{Prompt used for evaluating model responses in the medical reasoning task, with five human-aligned criteria and detailed instructions.}
\label{tab:medical-eval-prompt}
\begin{tabular}{p{0.97\linewidth}}
\toprule
\textbf{Medical Reasoning Task Evaluation Prompt} \\  
\midrule
Reference Information: \{reference\} \\

Answer to Score: \{answer\} \\

\vspace{0.5em}
Task: \\
Evaluate the given answer based on the provided reference information using the following criteria. Assign a score between 0 and 1 (inclusive) for each criterion, in increments of 0.1. A score of 1 means the answer fully meets the criterion, while a score of 0 means the answer fails to meet the criterion at all. \\

\vspace{0.5em}
\#\#\# Evaluation Criteria: \\

1. **Relevance** (Score: 0-1):  \\
   This criterion assesses how well the answer aligns with the reference information, addressing the symptoms, diagnosis, and treatments mentioned. The answer should directly respond to the medical context and conditions outlined in the reference. Answers that focus on the core issues presented, without deviating into irrelevant areas, should score higher. \\
   
\vspace{0.5em}
2. **Accuracy** (Score: 0-1):  \\
   The accuracy score reflects how correctly the answer represents the facts outlined in the reference. This includes correct medical terminology, diagnosis, and treatment recommendations. An answer should avoid introducing false or unsupported information while accurately reflecting the key aspects of the reference, including the medical procedures and conditions described. \\

\vspace{0.5em}
3. **Completeness** (Score: 0-1):   \\
   Completeness is assessed based on how thoroughly the answer covers the key points mentioned in the reference, including diagnostic procedures, symptoms, and treatment options. A complete answer should address all aspects of the medical condition mentioned in the reference, offering a full response to the query with relevant details. Missing important diagnostic tests or treatment steps will reduce the score. \\

\vspace{0.5em}
4. **Clarity** (Score: 0-1):   \\
   This criterion evaluates the clarity and readability of the answer. A high score is awarded to responses that are well-structured, logically coherent, and easily understood. An answer that communicates its reasoning in a clear and concise manner without ambiguity or unnecessary complexity will score higher. \\

\vspace{0.5em}
5. **Conciseness** (Score: 0-1):   \\
   This criterion evaluates how succinctly the answer conveys necessary information. Answers should avoid redundancy and irrelevant details but should not be penalized for adding depth and reasoning to the response. A longer, well-reasoned response that covers all necessary aspects of the reference will be rewarded, provided it does not become excessively verbose.  \\

\vspace{0.5em}
\#\#\# Response Format:   \\
Provide the evaluation results in the following format:  \\

\vspace{0.5em}
**Score Breakdown:**    \\

- **Relevance**: X.X (Explanation: [Provide brief reasoning for the score based on how well the answer aligns with the reference information and medical context]) \\

- **Accuracy**: X.X (Explanation: [Provide brief reasoning for the score based on the accuracy and consistency of the answer with the reference]) \\

- **Completeness**: X.X (Explanation: [Provide brief reasoning for the score based on the coverage of key points in the reference information]) \\ 

- **Clarity**: X.X (Explanation: [Provide brief reasoning for the score based on how clearly the answer is expressed]) \\

- **Conciseness**: X.X (Explanation: [Provide brief reasoning for the score based on how focused and concise the answer is]) \\

\bottomrule
\end{tabular}
\end{table}

\subsection{Reasoning Process Quality Metrics}
\label{appendix: reasoning quality metric}

In this section, we describe the metrics used to evaluate the quality of the reasoning process in the context of LLMs. Follow the previous work~\cite{zhou2025landscape}, we focus on three key metrics—\textbf{Consistency}, \textbf{Uncertainty}, and \textbf{Perplexity}—which are designed to measure the model’s reasoning stability and confidence during its multi-step reasoning process. These metrics are calculated on a set of sampled data and provide insights into the model's performance at different stages of reasoning.

\paragraph{Metric Definitions and Calculations.}
The three metrics used in this work are defined and calculated as follows:

1. \textbf{Consistency}:  
   Consistency measures the degree to which the model's reasoning remains stable over multiple reasoning steps. Specifically, we compute consistency by comparing the reasoning states at each step with the final state. If the model's reasoning process converges toward the correct answer, we expect higher consistency. The formula for consistency is:
   
    \begin{equation}
    \text{Consistency}(s_i) = \mathbf{I}(\arg\min s_i = \arg\min s_n)
    \label{eq:consistency}
    \end{equation}
   
   where \(s_i\) represents the reasoning state at the \(i\)-th step, and \(s_n\) is the final reasoning state. The indicator function \(\mathbb{I}\) outputs 1 if the states are identical (indicating convergence) and 0 otherwise.

2. \textbf{Uncertainty}:  
   Uncertainty quantifies how confident the model is about its predictions at intermediate steps. Higher uncertainty values indicate less confidence in the reasoning process. The uncertainty at a given step is calculated as the entropy of the state probabilities:
   
    \begin{equation}
    \text{Uncertainty}(s_i) = -\sum_{d \in s_i} d \cdot \log d
    \label{eq:uncertainty}
    \end{equation}
   
   where \(d\) represents the probability of a given state. This metric provides a measure of the model’s confidence in the reasoning path taken.

3. \textbf{Perplexity}:  
   Perplexity evaluates how well the model predicts the next token in the reasoning process, providing an indication of its confidence in the generated thoughts. Lower perplexity values correspond to more confident predictions. The formula for perplexity is:
   
    \begin{equation}
    \text{Perplexity}(t_i) = p_{\text{LLM}}(t_i | s_{i-1})^{-1 / |t_i|}
    \label{eq:perplexity}
    \end{equation}
   
   where \(t_i\) is the \(i\)-th token and \(s_{i-1}\) is the previous state. The calculation measures how likely the model is to generate the reasoning tokens at each stage, normalized by the token length.

\paragraph{Data Sampling and Metrics Calculation.}  
For evaluating the reasoning quality, we sampled 500 question-answer pairs from WebQSP and CWQ dataset. For each of these 500 QA pairs, we performed 10 independent inference runs using the model, generating 10 distinct reasoning paths per question. These repeated inferences allow for a more robust analysis of the reasoning process across multiple runs.

The reasoning paths were visualized for each case, and the consistency, uncertainty, and perplexity metrics were calculated for each of the 10 inference results per question. The metrics were then aggregated across the 500 samples to obtain an overall understanding of the reasoning behavior.

\paragraph{Case-Level Visualization.}  
To qualitatively understand the model's reasoning dynamics, we visualize each reasoning trajectory using landscape plots that depict the distribution of intermediate reasoning states. Following the method introduced in~\cite{zhou2025landscape}, the landscape is constructed by projecting the distance between intermediate thoughts and the final answer into a two-dimensional space. While the original work focuses on multiple-choice QA (where the distance is defined as perplexity between thoughts and answer options), we extend this formulation to open-ended QA by measuring the perplexity between thoughts and multi-answer targets. Each plot is divided into equal length ratio reasoning process segments, and density contours reflect the concentration of model-generated states. Darker regions indicate higher thought density, while green stars mark ground-truth answers. This allows us to track how the model's hypotheses evolve and converge, highlighting its alignment behavior and the effectiveness of symbolic reasoning during multi-hop inference.

\paragraph{Distribution-Level Analysis.}  
In addition to case-level visualizations, we also performed a statistical analysis over the 500 sampled QA pairs for WebQSP and CWQ. For the analysis, we calculated average of consistency, uncertainty, and median of perplexity across the reasoning steps. These values were then aggregated to analyze the overall trend of the model’s reasoning behavior across the entire set of questions. The results are presented as distributions, showing how these metrics change at each reasoning stage.

\section{Additional Results}

\subsection{Granular Performance Analysis}
In this section, we provide a more granular analysis of our model's performance on the WebQSP and CWQ test sets, broken down by the complexity of the questions. This analysis addresses the model's scalability on multi-hop reasoning and its robustness to questions with varying numbers of answers.

\subsubsection{Analysis by Hop Depth}
As shown in Table~\ref{tab:hop_analysis}, our model demonstrates strong performance across different hop depths. Notably, on the most complex >=3 hop questions in CWQ, KG-TRACES significantly outperforms prior SOTA methods, validating its effectiveness on deeper reasoning chains.

\begin{table*}[htbp]
\centering
\small
\caption{F1 score comparison with RoG for different question hops. Best results in each column are in bold.}
\label{tab:hop_analysis}
\begin{tabular}{l | ccc | ccc}
\toprule
\multirow{2}{*}{\textbf{Methods}} & \multicolumn{3}{c|}{\textbf{WebQSP}} & \multicolumn{3}{c}{\textbf{CWQ}} \\
\cmidrule(lr){2-4} \cmidrule(lr){5-7}
& 1 hop & 2 hop & $\ge 3$ hop & 1 hop & 2 hop & $\ge 3$ hop \\
\midrule
RoG & 77.03 & 64.86 & - & 62.88 & 58.46 & 37.82 \\
\textbf{KG-TRACES (ours)} & \textbf{80.53} & \textbf{67.23} & - & \textbf{64.26} & \textbf{58.73} & \textbf{45.37} \\
\bottomrule
\end{tabular}
\end{table*}

\subsubsection{Analysis by Answer Number}
Table~\ref{tab:ans_analysis} shows the model's performance on questions with different numbers of ground-truth answers. This demonstrates the model's robustness in handling both single-answer and multi-answer scenarios.

\begin{table*}[htbp]
\centering
\setlength{\tabcolsep}{5.3pt}
\scriptsize
\caption{F1 score comparison with RoG for questions with varying numbers of answers. Best results are in bold.}
\label{tab:ans_analysis}
\begin{tabular}{l | cccc | cccc}
\toprule
\multirow{2}{*}{\textbf{Methods}} & \multicolumn{4}{c|}{\textbf{WebQSP}} & \multicolumn{4}{c}{\textbf{CWQ}} \\
\cmidrule(lr){2-5} \cmidrule(lr){6-9}
& \#Ans=1 & 2$\ge$\#Ans$\le$4 & 5$\ge$\#Ans$\le$9 & \#Ans$\ge$10 & \#Ans=1 & 2$\ge$\#Ans$\le$4 & 5$\ge$\#Ans$\le$9 & \#Ans$\ge$10 \\
\midrule
RoG & 67.89 & 79.39 & 75.04 & \textbf{58.33} & 56.90 & \textbf{53.73} & 58.36 & 43.62 \\
\textbf{KG-TRACES (ours)} & \textbf{74.04} & \textbf{84.65} & \textbf{76.87} & 57.87 & \textbf{60.68} & 53.07 & \textbf{60.07} & \textbf{46.56} \\
\bottomrule
\end{tabular}
\end{table*}

\subsection{Path Retrieval Failure}
\label{app:fail}

We check whether at least one of the top-3 predicted relation paths can be traversed from the question entity in its retrieved subgraph. Table~\ref{tab:fail} reports the questions for which none of them can.

\begin{table}[h]
\centering
\caption{Test questions with no traversable path among the top-3 predicted relation paths.}
\label{tab:fail}
\begin{tabular}{lcc}
\toprule
Dataset & \#Test questions & No traversable path (top-3) \\
\midrule
WebQSP & 1,628 & 188 (11.5\%) \\
CWQ & 3,531 & 1,153 (32.7\%) \\
\bottomrule
\end{tabular}
\end{table}

Retrieval fails for 11.5\% of WebQSP questions and 32.7\% of CWQ questions. The failure rate is higher on CWQ, which contains more multi-hop questions.

\subsection{Attribution Accuracy}
To measure whether the provenance tags match the actual source of each claim, we use an LLM judge (GPT6-Sol) on 150 randomly sampled WebQSP and 150 CWQ test questions whose outputs contain both \texttt{<KG>} and \texttt{<INFERRED>} steps. The judge is given the question, the provided reasoning paths, and the generated reasoning with the tags removed, and decides whether each claim is supported by the provided paths.

\begin{table}[!h]
\centering
\caption{Support rate of claims under each provenance tag, judged against the provided KG paths.}
\label{tab:attr}
\begin{tabular}{lcc}
\toprule
Provenance tag & Supported / Total claims & Support rate \\
\midrule
\texttt{<KG>} & 1,414 / 1,745 & 81.0\% \\
\texttt{<INFERRED>} & 212 / 814 & 26.0\% \\
\bottomrule
\end{tabular}
\end{table}

Most \texttt{<KG>} claims (81.0\%) are supported by the provided paths, whereas most \texttt{<INFERRED>} claims are not, so the tags largely reflect the source of each claim.

\subsection{Reasoning Process Quality and Visualizations (Additional Results)}

\begin{figure}[hbtp]
    \centering
    \includegraphics[width=1\linewidth]{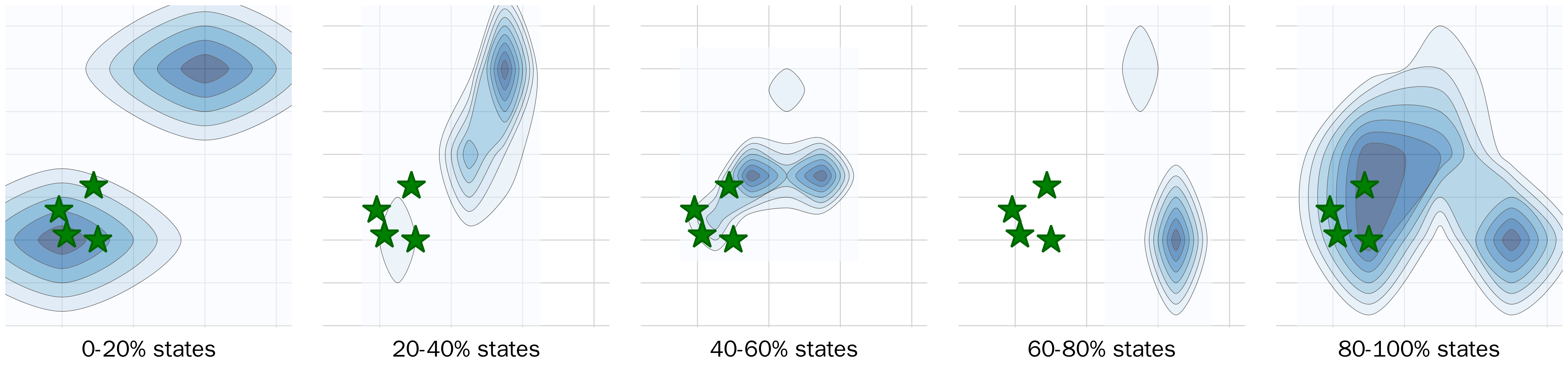}
    \caption{Visualization of model reasoning thoughts for a representative case 1 in WebQSP}
    \label{fig:webqsp-case-1-reasoning-trajectory}
\end{figure}

\begin{figure}[hbtp]
    \centering
    \includegraphics[width=1\linewidth]{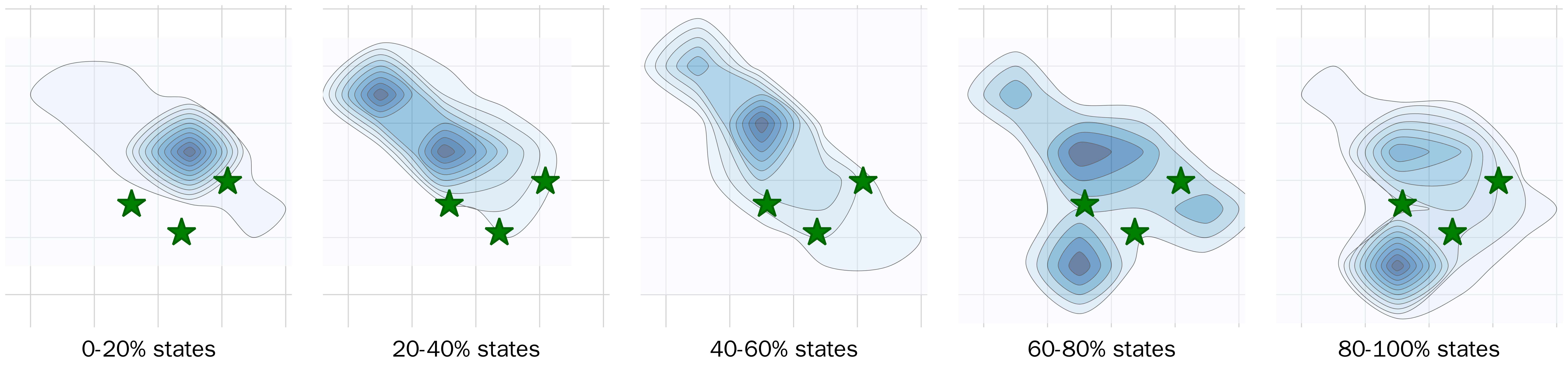}
    \caption{Visualization of model reasoning thoughts for a representative case 2 in WebQSP}
    \label{fig:webqsp-case-2-reasoning-trajectory}
\end{figure}

\begin{figure}[hbtp]
    \centering
    \includegraphics[width=1\linewidth]{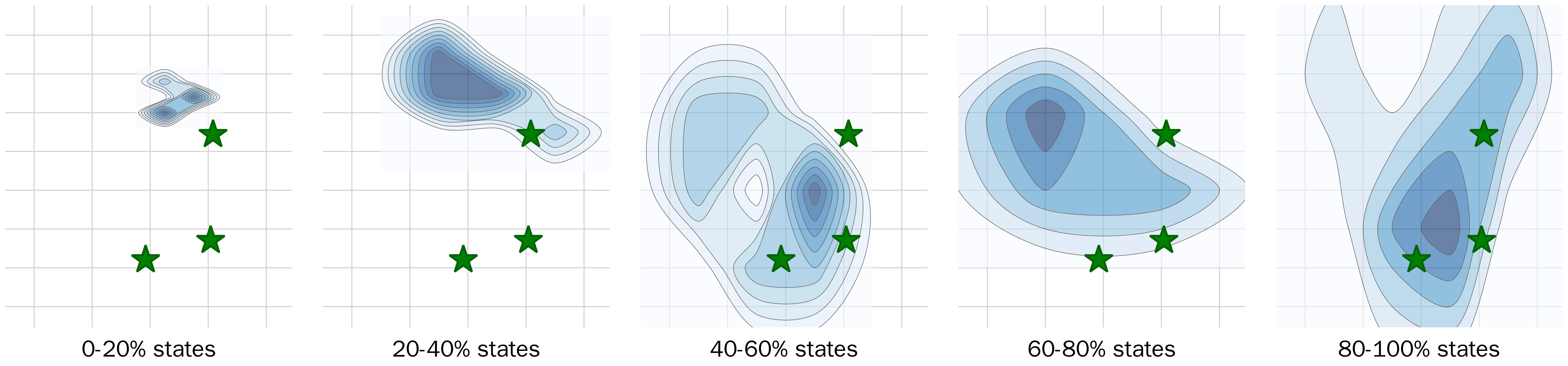}
    \caption{Visualization of model reasoning thoughts for a representative case 3 in WebQSP}
    \label{fig:webqsp-case-3-reasoning-trajectory}
\end{figure}

\begin{figure}[H]
    \centering
    \includegraphics[width=1\linewidth]{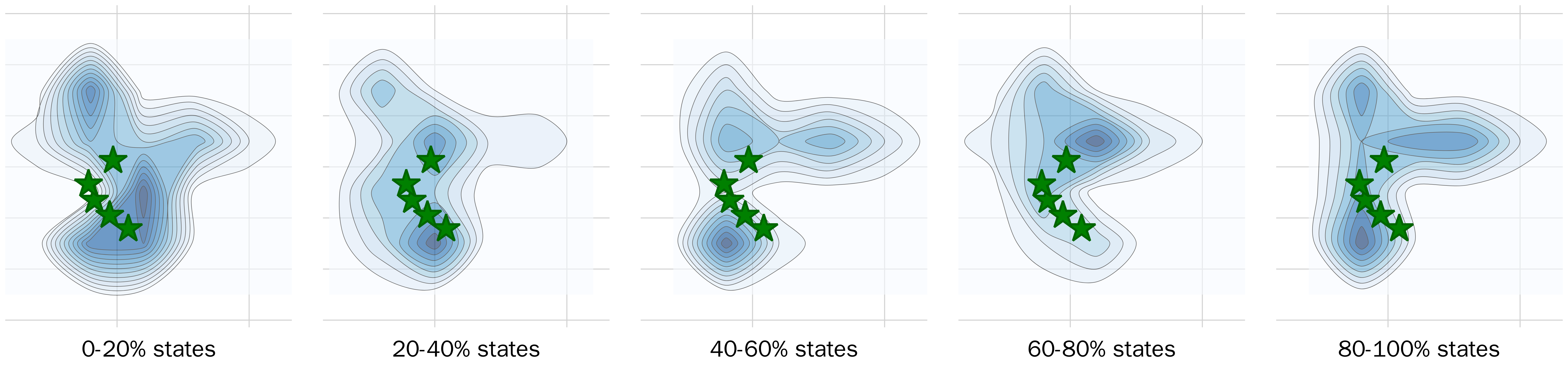}
    \caption{Visualization of model reasoning thoughts for a representative case 1 in CWQ}
    \label{fig:cwq-case-1-reasoning-trajectory}
\end{figure}

\begin{figure}[!t]
    \centering
    \includegraphics[width=1\linewidth]{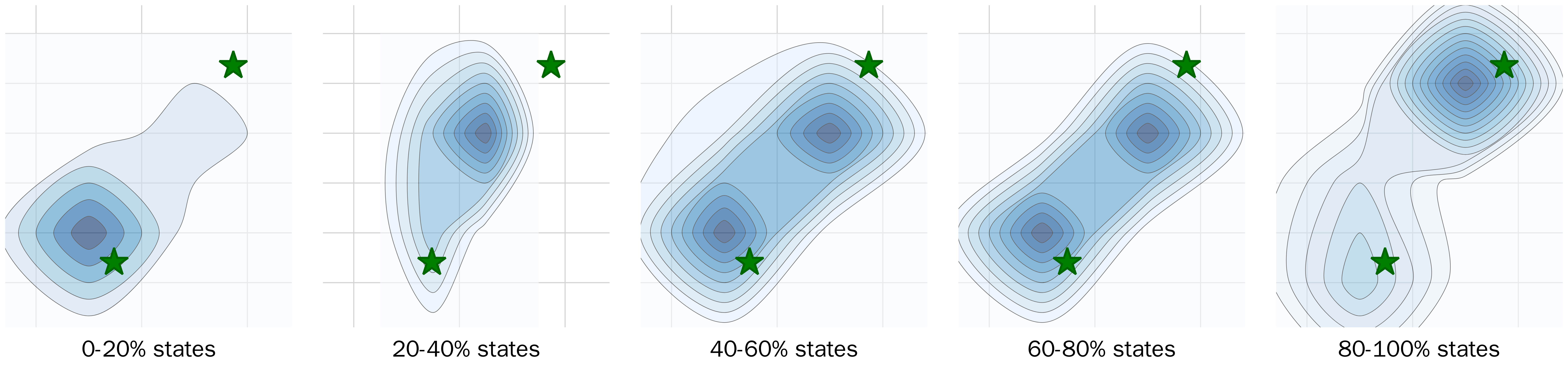}
    \caption{Visualization of model reasoning thoughts for a representative case 2 in CWQ}
    \label{fig:cwq-case-2-reasoning-trajectory}
\end{figure}

\begin{figure}[H]
    \centering
    \includegraphics[width=1\linewidth]{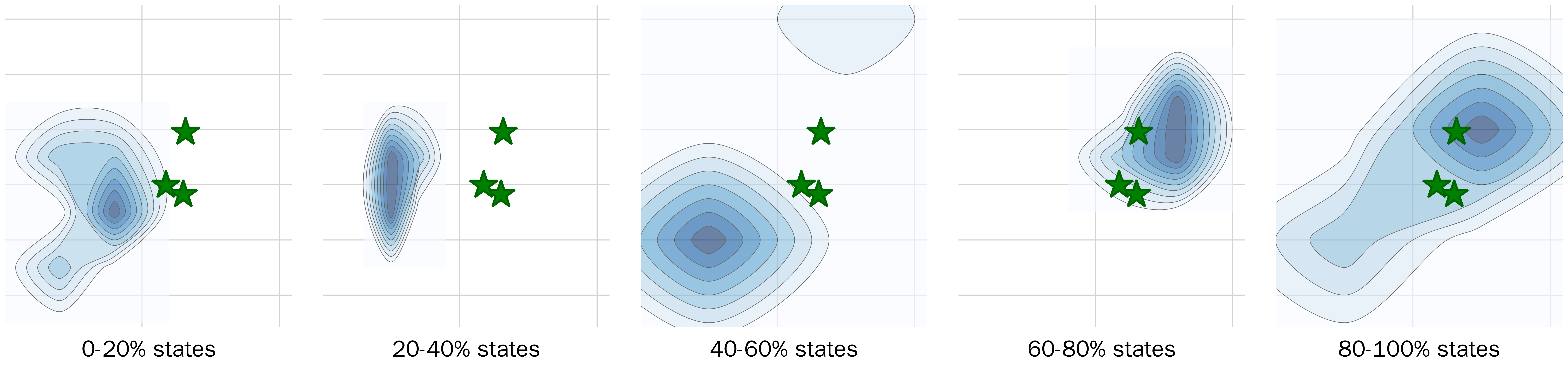}
    \caption{Visualization of model reasoning thoughts for a representative case 3 in CWQ}
    \label{fig:cwq-case-3-reasoning-trajectory}
\end{figure}

As Figure~\ref{fig:webqsp-case-1-reasoning-trajectory}, Figure~\ref{fig:webqsp-case-2-reasoning-trajectory}, Figure~\ref{fig:webqsp-case-3-reasoning-trajectory}, Figure~\ref{fig:cwq-case-1-reasoning-trajectory}, Figure~\ref{fig:cwq-case-2-reasoning-trajectory}, and Figure~\ref{fig:cwq-case-3-reasoning-trajectory} show, we provide other 3 reasoning process visualization case of KG-TRACES in WebQSP and CWQ respectively.

\end{document}